\documentclass[10pt,twocolumn,letterpaper]{article}

\usepackage{iccv}              

\usepackage{natbib}
\usepackage{graphicx}

\usepackage{amsmath,amsfonts,bm}

\def\eqref#1{equation~\ref{#1}}

\def\1{\bm{1}}

\DeclareMathAlphabet{\mathsfit}{\encodingdefault}{\sfdefault}{m}{sl}
\SetMathAlphabet{\mathsfit}{bold}{\encodingdefault}{\sfdefault}{bx}{n}

\usepackage{xcolor}

\usepackage{url}
\usepackage{xspace}
\usepackage{caption}
\usepackage{multirow}
\usepackage{multicol}

\newcommand{\methodname}[0]{REM\xspace }

\newcommand{\smallsec}[1]{\vspace{0.2em}\noindent\textbf{#1}}

\definecolor{iccvblue}{rgb}{0.21,0.49,0.74}
\usepackage[pagebackref,breaklinks,colorlinks,allcolors=iccvblue,urlcolor=magenta]{hyperref}
\def\paperID{3615} 
\def\confName{ICCV}
\def\confYear{2025}

\title{ReferEverything: Towards Segmenting Everything We Can Speak of in Videos}

\author{
Anurag Bagchi$^{1}$ \quad 
Zhipeng Bao$^{1}$ \quad 
Yu-Xiong Wang$^{2}$ \quad 
Pavel Tokmakov$^{3}$\textsuperscript{\dag} \quad 
Martial Hebert$^{1}$\textsuperscript{\dag} \\
{\normalsize $^{1}$Carnegie Mellon University \quad $^{2}$University of Illinois Urbana-Champaign \quad $^{3}$Toyota Research Institute} \\
\vspace{-0.4cm} \\
{\normalsize \url{https://refereverything.github.io/}}
}

\begin{document}

\twocolumn[{%
    \maketitle
    \begin{center}
        \centering
        \vspace{-20pt}
        \includegraphics[width= \linewidth]{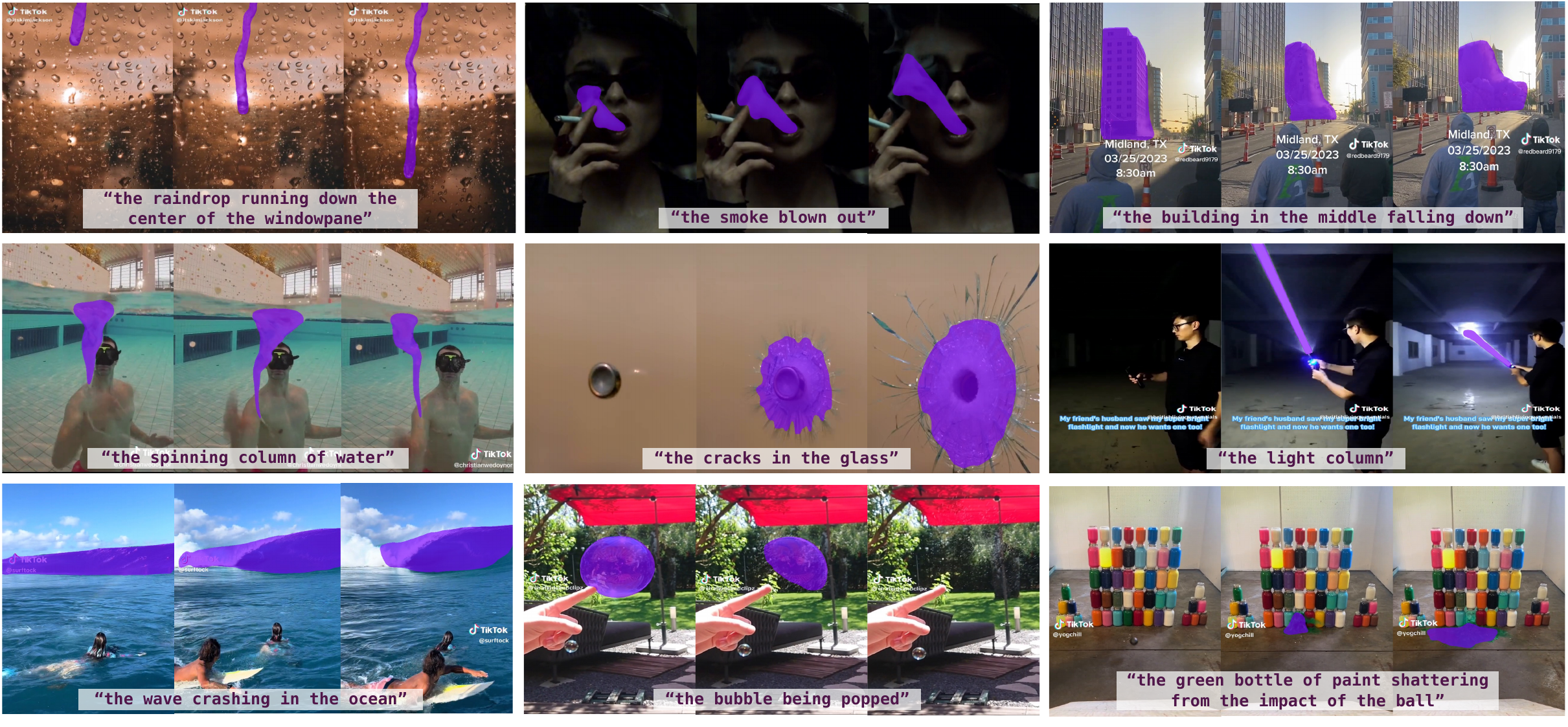}
        \vspace{-20pt}
        \captionof{figure}{Results of our method, \methodname, which leverages visual-language representations learned by video diffusion models to segment a wide range of concepts based on natural language descriptions (shown in grey boxes). \methodname exhibits zero-shot generalization to challenging, dynamic concepts, such as raindrops or shattering glass. Video visualizations are available on the \href{https://refereverything.github.io/\#REM}{project page}.}
        \label{fig:teaser}
    \end{center}
}]

\begingroup
\renewcommand\thefootnote{\dag}
\footnotetext{Equal advising.}
\endgroup

\begin{abstract}
We present \methodname, a framework for segmenting a wide range of concepts in video that can be described through natural language. Our method leverages the universal visual-language mapping learned by video diffusion models on Internet-scale data by fine-tuning them on small-scale Referring Object Segmentation datasets. Our key insight is to preserve the entirety of the generative model's architecture by shifting its objective from predicting noise to predicting mask latents. The resulting model can accurately segment rare and unseen objects, despite only being trained on a limited set of categories. Additionally, it can effortlessly generalize to non-object dynamic concepts, such as smoke or raindrops, as demonstrated in our new benchmark for Referring Video Process Segmentation (Ref-VPS).
\methodname performs on par with the state-of-the-art on in-domain datasets, like Ref-DAVIS, while outperforming them by up to 12 IoU points out-of-domain, leveraging the power of generative pre-training. We also show that advancements in video generation directly improve segmentation.
\end{abstract}


\section{Introduction}
\label{sec:introduction}

\begin{figure*}[t]
    \centering
    \includegraphics[width= \linewidth]{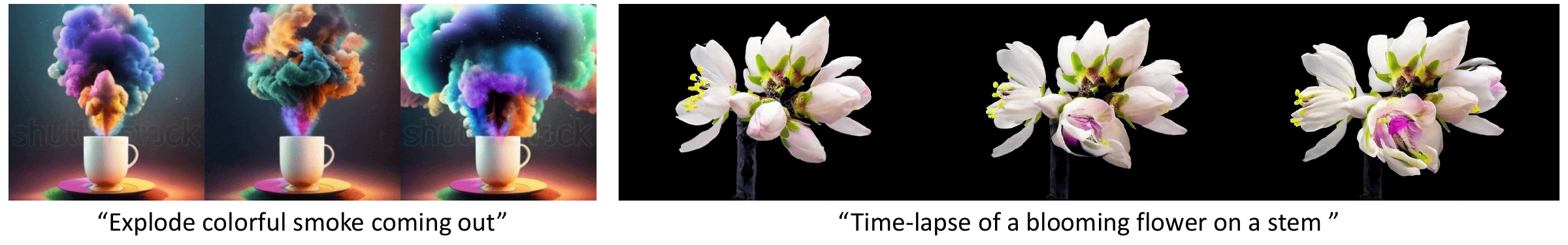}
    \vspace{-20pt}
    \caption{Through Internet-scale pre-training, video diffusion models learn to generate videos capturing the diversity of the dynamic visual world (samples shown above). We leverage their powerful visual-language representation for open-world referring video segmentation.}
    \label{fig:motivation}
    \vspace{-12pt}
\end{figure*}

One of the most remarkable properties of natural language lies in its ability to convey the richness and complexity of human visual experience. From fleeting moments, like raindrops rolling down the window or smoke dissipating from a cigarette (see row 1 in Figure~\ref{fig:teaser}), to dynamic processes, such as a glass shattering or a whirlpool forming in the water (row 2 in Figure~\ref{fig:teaser}), language enables us to descrive and reference events with precision. Crucially, if an event can be expressed verbally, it can often be accurately localized in both space and time. This universal mapping between the discrete, symbolic realm of language and the continuous, ever-changing visual world is developed through a lifetime of visual-linguistic interaction~\citep{barsalou1999perceptual,popham2021visual}.

The corresponding task in computer vision - Referring Video Segmentation (RVS)~\citep{gavrilyuk2018actor, hu2016segmentation}, is defined as the task of segmenting a specific region in a video based on a natural language expression. 
However, virtually all existing literature focuses on a narrow subset of RVS - Referring Video \textit{Object} Segmentation (RVOS)~\citep{seo2020urvos,wu2022language}. This emphasis stems from data constraints: RVOS datasets were built by annotating object tracking benchmarks~\citep{davis2017, xu2018youtube}, which are inherently object-centric and limited in scale. Recent advances in large visual-language datasets~\citep{schuhmann2022laion, bain2021frozen} and generative models~\citep{rombach2022high, wang2023modelscope,wan2025wan} present opportunities to overcome these limitations. Visual-language representations learned by these models from Internet data demonstrated strong generalization in image-based segmentation~\citep{vpd23, ozguroglu2024pix2gestalt}, but their potential in videos remains underexplored~\citep{zhu2024exploring}.  

In this work, we introduce \methodname~-- a novel approach to RVS that enables spatio-temporal localization of a wide range of concepts~\citep{ghorbani2019towards} in video that can be described through natural language, beyond conventional object tracking (shown in Figure~\ref{fig:teaser}). A key factor behind \methodname's success is preserving the universal visual-language mapping learned by generative models during Web-scale pre-training (see Figure~\ref{fig:motivation}). To this end, we retain the original model architecture and fine-tune it on small-scale RVOS datasets, adjusting the output to generate target mask latents instead of Gaussian noise. As shown in Table~\ref{tab:rvos}, this model, originally designed and trained for video generation, demonstrates competitive performance against specialized RVOS methods on popular benchmarks~\citep{davis2017, xu2018youtube}. More significantly, it exhibits much stronger generalization.

To quantify this effect, we report \textit{zero-shot} evaluation results on the open-world object tracking benchmark - BURST~\citep{athar2023burst}, and the non-object `Stuff' segmentation dataset - VSPW~\citep{miao2021vspw}, as well as on a newly collected benchmark that expands the focus of RVS to include dynamic process in Section~\ref{sec:dataset}. We define the latter as temporally evolving \textit{events}, where subjects undergo continuous changes in state, shape, or appearance (see examples in Figure~\ref{fig:teaser}). 
Our new benchmark, which we call Ref-VPS for Referring Video Process Segmentation, consists of 145 videos that are labeled with referring expressions and masks at 6 fps and span 39 unique concepts. Experiments in Section~\ref{sec:experiment_refproc} demonstrate that existing approaches, including the very recent method of \citet{zhu2024exploring}, fail to generalize outside of the narrow training distribution, whereas our method effortlessly segments a wide spectrum of targets (see Figures~\ref{fig:teaser} and~\ref{fig:rvos}). 

The primary contribution of this paper is to demonstrate that Web-scale video diffusion models have learned universal visual-language mapping that can be repurposed for open-world referring video segmentation. We further introduce a new benchmark for Referring Video Process Segmentation in Section~\ref{sec:dataset},  expanding the focus of RVS beyond conventional object tracking. Finally, we provide a detailed analysis of our approach in Section~\ref{sec:abl}, demonstrating that retaining the full architecture of the generative model, rather than isolating the de-noising network as a feature extractor, is key to unlocking the strongest generalization in RVS. 

\section{Related Work}
\label{sec:rel_work}

\smallsec{Referring Video Segmentation (RVS)} involves segmenting specific regions in a video based on a natural language description~\citep{gavrilyuk2018actor, khoreva2019video, seo2020urvos}. Most benchmarks for this task were developed by adding referring expression annotations to existing Video Object Segmentation (VOS) datasets, such as DAVIS'17~\citep{davis2017} or YouTube-VOS~\citep{xu2018youtube}. Consequently, the role of language in these benchmarks is limited to providing an interface for user-initialized object tracking~\citep{wu2013online,perazzi2016benchmark}. While segmenting objects is valuable, it addresses only a narrow subset of the possible interactions between language and the space-time continuum of videos.
Equally important is the ability of RVS methods to segment video concepts beyond common object categories.
To address this gap, we introduce a new benchmark focused on segmenting dynamic processes, which we term Referring Video Process Segmentation (Ref-VPS).

Earlier RVOS approaches~\citep{bellver2020refvos,ning2020polar,hui2021collaborative} generally employed a \emph{bottom-up} strategy: first, image-level methods~\citep{rother2004grabcut,ye2019cross,carion2020end,plummer2015flickr30k} were applied to obtain frame masks, followed by spatio-temporal reasoning, such as mask propagation~\citep{seo2020urvos}, to refine the segmentation across frames. More recently, with the success of cross-attention-based methods ~\citep{vaswani2017attention,meinhardt2022trackformer,zeng2022motr} in object segmentation and tracking, query-based architectures have been introduced to RVOS, leading to significant improvements~\citep{wu2022language,wu2023onlinerefer, yan2024referred}. The limited scale of paired video-language data with segmentation annotations has always been a major limitation, causing most methods to train jointly on videos and images~\citep{kazemzadeh2014referitgame,jhuang2013towards}. The latest approaches go further and unify all object localization tasks in a single framework~\citep{yan2023universal,wu2024general,cheng2023tracking}. However, while these models excel in object tracking, they struggle to generalize to more dynamic concepts. In contrast, we demonstrate that generative pre-training on Internet-scale data~\citep{schuhmann2022laion,bain2021frozen} results in a universal (\ie, not limited to one domain) mapping between the space of language and the ever-changing visual world.

\smallsec{Diffusion models} have emerged as the de facto standard for generative learning in computer vision~\citep{sohl2015deep,ho2020denoising} and beyond~\citep{chi2023diffusion}. Among them, the Denoising Diffusion Probabilistic Model (DDPM)~\citep{ho2020denoising} leverages neural network components to model the denoising process. Stable Diffusion (SD)~\citep{rombach2022high} shifts the denoising process into the latent space of a pre-trained autoencoder~\citep{kingma2013auto}, allowing for model scaling. Expanding from images to videos, diffusion models have seen success in text-to-video (T2V) generation~\citep{wang2023modelscope,chen2023videocrafter1,chen2024videocrafter2,opensora,blattmann2023stable}. In addition to the capacity to generate high-fidelity images based on text prompts, the T2V diffusion models implicitly learn the mapping from linguistic descriptions to video regions, providing an opportunity to repurpose them for RVOS. Among current T2V methods, ModelScope~\citep{wang2023modelscope} and Wan~\citep{wan2025wan} stand out for their open-source implementations and top performance.

\smallsec{Visual-language pre-training for perception:} in addition to being highly effective in image and video generation, diffusion models have been shown to learn a strong representation of the natural image manifold. Several works demonstrated that they can be re-purposed for computer vision problems, including semantic segmentation~\citep{xu2023odise,vpd23,zhang2024tale} and pixel-level correspondence~\citep{tang2024emergent}, achieving an impressive degree of generalization. Others have shown that image diffusion models learn powerful object representations, enabling open-world novel view synthesis~\citep{liu2023zero} and amodal segmentation~\citep{ozguroglu2024pix2gestalt}.
Most recently, \citet{zhu2024exploring} also leverages pre-trained T2V models for RVOS, however, our analysis shows that their approach fails to fully capitalize on the universal visual-language mapping learned in generative pre-training. 
In this work, we explore the application of video diffusion models to RVS, 
demonstrating how to maintain a high-level generalizability during fine-tuning.

In a separate line of work, visual-language representations learned with contrastive objectives~\citep{bao2022generative,radford2021learning} have been adapted for referring image~\citep{lai2024lisa,rasheed2024glamm,you2023ferret,xu2024pixel} and video segmentation~\citep{zhou2024driving}. However, their performance remains limited compared to both generative models and classical referring segmentation approaches.

\begin{figure*}[t]
    \centering
    \includegraphics[width=0.9\linewidth]{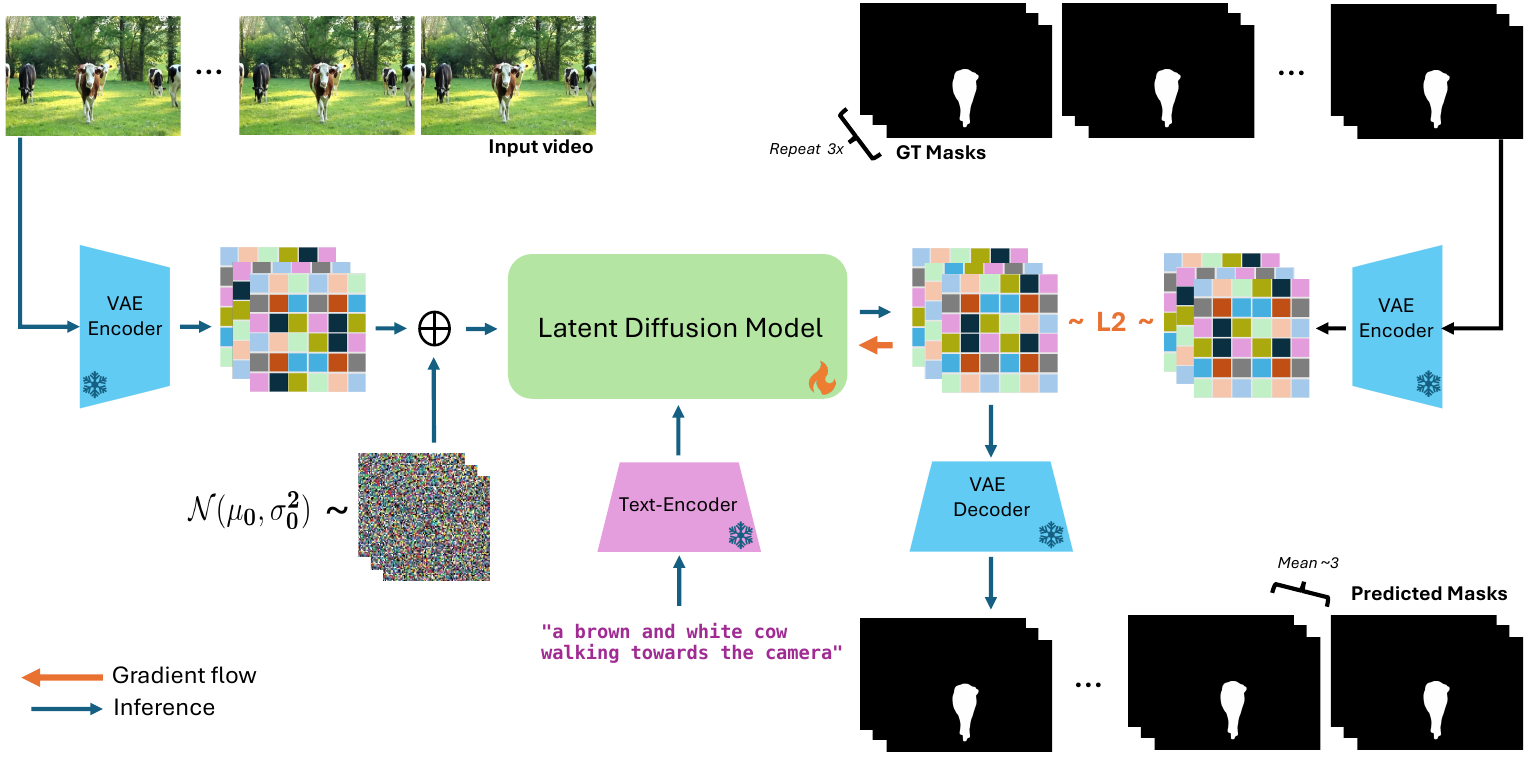}
    \caption{The model architecture of Refer Everything with Diffusion Models (\methodname). Like a video diffusion model it is based on, our approach takes video frames with added noise and a language expression as input. Our key insight is preserving the entirety of the generative model's architecture by shifting its objective from predicting noise to predicting mask latents.}
    \label{fig:method}
    \vspace{-12pt}
\end{figure*}
\section{Method}
\label{sec:method}

\subsection{Visual-language mapping via video denoising}

Text-to-Video (T2V) diffusion models~\citep{wang2023modelscope,chen2024videocrafter2,opensora} generate videos that align with a given language description, starting from Gaussian noise. The process can be formalized as:
\begin{equation}
    \hat{x} = f_\text{vdm} (x_T, c, T),
    \label{eq:vdmsetting}
\end{equation}
where $\hat{x}$ is the generated video, $T$ denotes the maximum timestep specified by the video diffusion model $f_\text{vdm}$, $x_T$ is a sample drawn from a Gaussian distribution $\mathcal{N}(\mu_T, \sigma_T^2)$ predefined by the video diffusion model, and $c$ is the conditioning prompt. To reduce computational complexity, these models often perform denoising in the latent space~\citep{rombach2022high}. Specifically, a pre-trained Variational Autoencoder (VAE)~\citep{kingma2013auto} is employed to map the video $x$ from pixel space into latent space, denoted as $\mathcal{E}(x) = z$, while a decoder reconstructs it from the latent space, $\mathcal{D}(z) \approx x$. Thus, the generation process becomes:
\begin{equation}
    \hat{x} = \mathcal{D}\left ( f_\text{vdm} (z_T,c,T) \right).
\end{equation}
During training, rather than denoising from pure Gaussian latents, T2V models denoise from partially noisy video latents and optimize the following latent diffusion objective:
\begin{equation}
    \min_\theta \mathbb{E}_{z\sim \mathcal{E}(x), t, \epsilon \sim \mathcal{N}  (0,1)}\left \| \epsilon - \epsilon_\theta(z_t, e_c, t)\right \|_2^2  ,
    \label{eq:denoise}
\end{equation}
where $\epsilon$ is the Gaussian noise added to the clean video latent, $z_t$ represents the noisy video latent at timestep $t$ derived by the diffusion forward pass~\citep{ho2020denoising,rombach2022high}, and $e_c$ is the conditional embedding generated from $c$ using a text encoder~\citep{radford2021learning}. The denoising network $\epsilon_\theta(z_t, e_c, t)$, typically a U-Net~\citep{ronneberger2015u} or a Diffusion Transformer (DiT)~\cite{peebles2023scalable}, is tasked with predicting the noise $\epsilon$. In this network, the conditional embedding $e_c$ interacts with the latent representations through attention mechanisms, guiding the model to generate diverse, semantically accurate videos.

\subsection{From language-conditioned denoising to RVS}
\label{sec:ourmethod}

Referring Video Segmentation (RVS) involves segmenting an entity in a video across spatial and temporal dimensions, guided by a natural language description. Formally, the task is defined as:
\begin{equation}
    \hat{m} = f_\text{RVS}(x, c),
\end{equation}
where $f_\text{RVS}$ is the RVS model, $x$ represents a video sequence, $c$ denotes a referring text prompt, and $\hat{m}$ is the binary masks produced as output. This task aligns naturally with T2V models, which establish a robust mapping between the entities described in the text and the corresponding spatial-temporal regions in the video by optimizing the denoising objective in Equation~\ref{eq:denoise}.

Several prior works have explored the alignment of diffusion models with referring segmentation~\citep{vpd23,xu2023odise,zhu2024exploring}, typically employing these models as \textit{feature extractors}. Specifically, they adjust the input format of a referring segmentation model to match that of the denoising network $\epsilon_\theta$, and pass the resulting features to a task-specific decoder $f_\text{dec}$ (\eg, a convolutional network) to predict the target masks:
\begin{equation}
    \hat{m} = f_\text{dec}(\epsilon_\theta^{(n)}(z_t, e_c, t)),
\end{equation}
where $z_t$ is the noisy latent representation of the input images at timestep $t$, $e_c$ is a feature embedding of the referring expression $c$, and $\epsilon_\theta^{(n)}$ denotes the intermediate feature at the $n^{th}$ layer. In practice, $t$ is usually set to a small value (\eg, 50), and $n$ is set to the later layer indexes to obtain the optimal performance. The entire model is then trained in a conventional discriminative learning setup. 
However, replacing parts of the generative model with newly initialized layers can disrupt alignment between the model's representation from pre-training and the new features learned on narrow-domain datasets, leading to a substantial loss of generalization capabilities.

In our approach, shown in Figure~\ref{fig:method}, we propose to instead preserve the architecture of the video diffusion model in its entirety. Specifically, rather than using intermediate features $\epsilon_\theta^{(n)}$, \methodname repurposes the whole denoising network $\epsilon_\theta$ (together with the VAE) by shifting its objective from predicting noise to predicting mask \textit{latents} (shown on the right in Figure~\ref{fig:method}): 
\begin{equation}
    \hat{m} = \mathcal{D}(\epsilon_\theta(z_t, e_c, t)),
    \label{eq:rvsinfernece}
\end{equation}
where $\mathcal{D}$ denotes the (frozen) VAE decoder used to produce the actual binary segmentation masks from the predicted latents. That is, instead of learning the decoder network $f_\text{dec}$ from scratch, we reuse the VAE from the video diffusion model.
This subtle yet powerful modification allows the model to better preserve its universal visual-language mapping learned on Internet-scale data during generative pre-training while adapting to the task of RVS.

\smallsec{Training and optimization.} During training, to encode the ground-truth segmentation masks with the VAE, we broadcast the single-channel mask into three channels by simply duplicating it (shown in the top right of Figure~\ref{fig:method}). For simplicity, we still denote this three-channel mask representation as $m$. The pre-trained VAE can then map the mask sequence into the latent space via $\mathcal{E}(m) = z^m$ and decode the masks back from predicted latents via $\mathcal{D}(z^m) \approx m$. 
For the noisy latent $z_t$ and timestep $t$, we prioritize using latents that remain as clean as possible. Therefore, we always set the timestep to its minimum value, $t=0$. To train the model, we supervise the predicted mask latents using an $\mathcal{L}_2$ loss (shown in the center-right of Figure~\ref{fig:method}) by minimizing:
\begin{equation}
\min_\theta \mathbb{E}_{z^m\sim \mathcal{E}(m), t=0}\left \| z^m - \mathbf{\epsilon}_\theta(z_t, e_c, t) \right \|_2^2.
\end{equation}

\smallsec{Model inference.} During inference, we follow Equation~\ref{eq:rvsinfernece}, with $t=0$, to decode the predicted mask latent and generate three-channel mask predictions. We then compute the single-channel masks by averaging the pixel values of the three channels and applying a constant threshold of 0.5 to binarize the result (shown in the bottom right of Figure~\ref{fig:method}). Notably, the inference is non-iterative (the mask is predicted in a single forward pass), making the computational cost of \methodname on par with other approaches in the literature.

\section{Benchmark Design and Collection}
\label{sec:dataset}
Existing datasets only allow for quantifying the generalization of RVS models to rare \textit{object} categories~\citep{athar2023burst} or \textit{static} `Stuff'~\citep{miao2021vspw}. 
However, covering the entire spectrum of concepts that can be spoken of in videos in a single benchmark would be extremely costly. Instead, we narrow our focus to the most salient subset that necessitates joint modeling of language and temporal dynamics. Specifically, we target \textbf{dynamic processes}, defined as temporally evolving \textit{events}, where the subjects undergo continuous changes in state, shape, or appearance. Importantly, these subjects are not limited to objects but include any spatio-temporally localizable phenomena, such as light or fire. Our new benchmark, Referring Video Process Segmentation (Ref-VPS), is built by selecting representative videos and annotating them with referring expressions and segmentation masks.

\subsection{Video selection}
To source the videos, we require a large, public, and diverse database that supports natural language queries and permits content redistribution for research. Based on these criteria, we selected \href{https://www.tiktok.com}{TikTok} — a platform with 
tens of millions of daily uploads that capture a wide range of dynamic scenarios. Moreover, TikTok’s policies generally allow free redistribution of content, with individual users retaining the option to opt out.

To identify a representative pool of videos, we established a taxonomy of five broad, possibly overlapping concepts (\eg, `object transformations', or `pattern evolution'; see Section A in the supplementary for definitions). Although these do not cover every possible dynamic process, we selected them to create a clear, focused framework for sourcing representative samples. For each concept, we use ChatGPT~\citep{openai2023chatgpt} to generate concrete examples and search queries for TikTok (\eg, `a wax candle melting' for `object transformations'), resulting in 120 fine-grained concepts. The queries retrieved over 1,000 samples, but many were unsuitable - either because the events (\eg, `glaciers melting over time') occur over extended periods and are rarely captured on TikTok, or because of ambiguous search terms. After this step, we obtained a set of 342 samples.

We then manually filtered the videos using these criteria: (1) exclude videos lacking significant dynamic changes (\eg, mostly stationary clouds); (2) exclude events that occur too rapidly to label a sufficient number of non-empty frames (\eg, lightning flashes); and (3) exclude videos with frequent shot changes that prevent extraction of a continuous clip capturing the event. Additionally, for videos that are compilations of similar events, we split them into individual clips and treat each independently. The final dataset comprises 145 clips covering 39 concepts. As it is intended for zero-shot evaluation, no additional splits are defined. Representative samples with results are shown in Figure~\ref{fig:teaser}.

\subsection{Annotation collection and evaluation}
\begin{figure}[t]
    \centering
    \includegraphics[width=\linewidth]{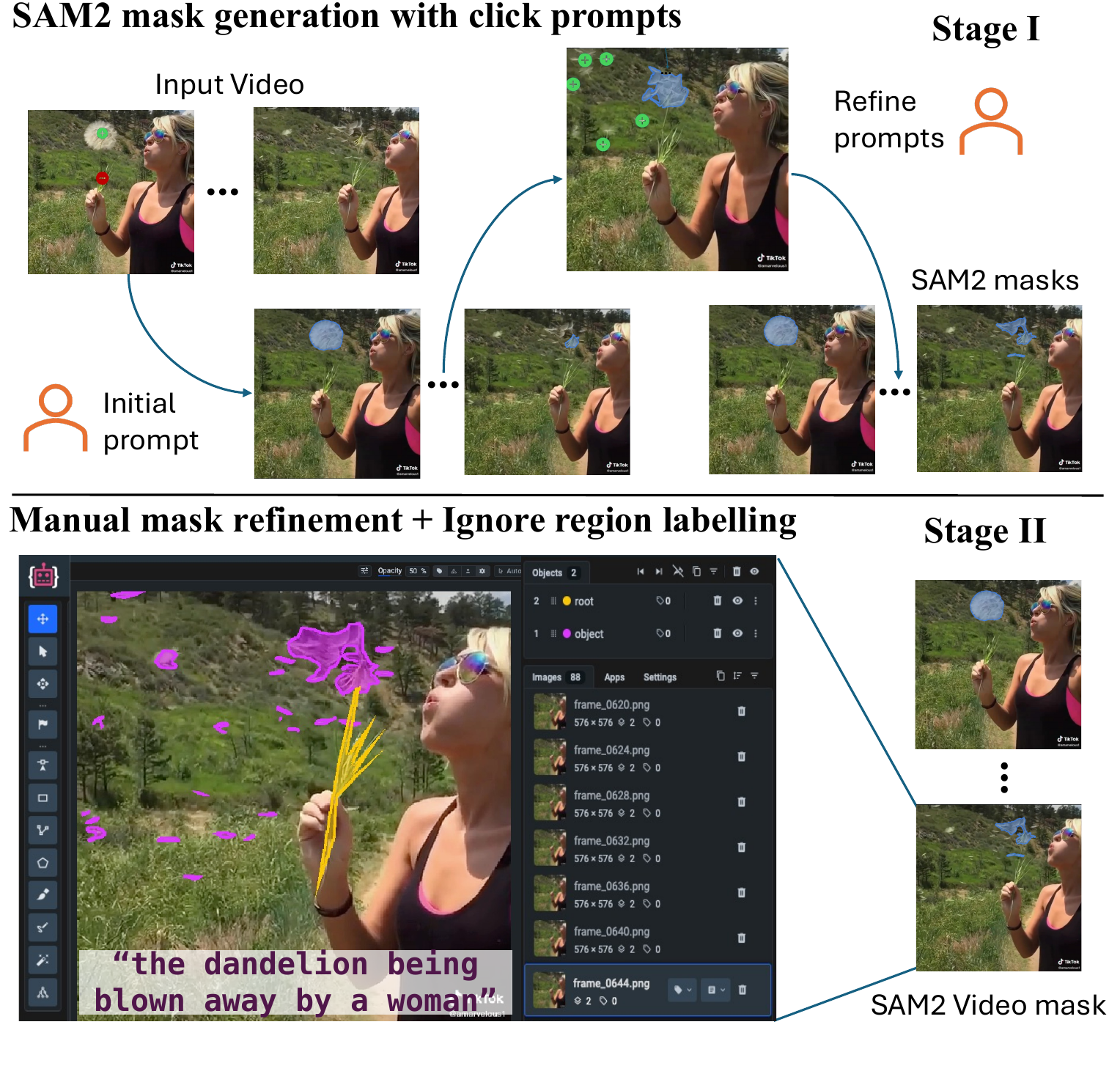}
    \vspace{-20pt}
    \caption{Our mask annotation pipeline. We first use SAM2 to interactively segment the region of interest in the video (shown above). We then manually refine the masks where SAM2 fails and label ambiguous regions as Ignore (shown in yellow below).}
    \label{fig:annotation_process}
    \vspace{-10pt}
\end{figure}
To label the selected videos, we first adjust each clip's temporal boundaries to focus on the event of interest and avoid shot changes, ensuring that the event, along with some contextual frames before and after, is captured in its entirety. The clips are then exported as frames at 24 FPS.

To collect referring expressions, we first manually identify the target entity in each clip and then instruct two independent annotators to provide descriptions. Each annotator contributes two expressions per target, yielding a total of four distinct expressions per clip. Following standard protocol~\citep{khoreva2019video, seo2020urvos}, evaluation is conducted on all queries, and results are reported as the average performance across them.

Next, we label the identified targets with segmentation masks at 6 FPS using a semi-automatic pipeline. To this end, we leverage the recently introduced SAM2~\citep{ravi2024sam} foundational model for interactive video segmentation. In particular, as shown in the top part of Figure~\ref{fig:annotation_process}, we first provide positive and negative clicks to a frame of the video. SAM2 then automatically segments the entity of interest in the frame, as well as propagates the mask across the entire clip. We interactively improve segmentation quality by providing additional clicks as needed. 

\begin{table*}[t]
    \centering
    \resizebox{\linewidth}{!}{
    \begin{tabular}{l|cc|ccc|ccc}
         \multirow{2}{*}{Method} &  \multirow{2}{*}{Pre-training Data} &  \multirow{2}{*}{Mask/Box Supervision} &  \multicolumn{3}{c|}{Ref-DAVIS}   &  \multicolumn{3}{c}{Ref-YTB}  \\ 
         \cline{4-9} 
        & & & $\mathcal{J} \& \mathcal{F}$  & $\mathcal{J}$ & $\mathcal{F}$ & $\mathcal{J} \& \mathcal{F}$  & $\mathcal{J}$ & $\mathcal{F}$  \\
        \hline
         Referformer~\citep{wu2022language} & ImageNet + Kinetics + SSv2 & Ref-COCO/+/g + Ref-YTB  & 61.1 & 58.1 & 64.1 & 62.9 & 61.3 & 64.6 \\
         MUTR~\citep{yan2024referred} & ImageNet + Kinetics + SSv2& Ref-YTB + AVS & 68.0 & 64.8 & 71.3 & 68.4 & 66.4 & 70.4 \\
         VLMO-L~\citep{zhou2024driving} & Unknown & Ref-COCO/+/g + Ref-YTB  & 70.2 & 66.3 & 74.1 & 67.6 & 65.3 & 69.8\\
         UNINEXT~\citep{yan2023universal} & Object365& 10+ Image/Video datasets  & 72.5 & 68.2 & 76.8 & 70.1 & 67.6 & 72.7 \\
         VD-IT~\citep{zhu2024exploring} & LAION5B + WebVid &  Ref-COCO/+/g + Ref-YTB  & 69.4 & 66.2 & 72.6 & 66.5 & 64.4 & 68.5 \\
         \hline
         \methodname (MS-1.4B) & LAION5B + WebVid & Ref-COCO/+/g + Ref-YTB  & 72.6 & 69.9 & 75.3 & 68.4 & 67.1 & 69.7\\
         \methodname (Wan-14B) & Internal + Public Images/Videos & Ref-COCO/+/g + Ref-YTB  & \bf 75.0 & \bf 71.3 & \bf 78.7 & \bf 71.7 & \bf 69.2 & \bf 74.3\\
         
    \end{tabular}
    }
    \caption{Comparison to the state of the art on the validation set of the Ref-DAVIS and the test set of Ref-YTB benchmarks using the standard metrics. Even the base version of our method performs on par with the strong UNINEXT approach, despite not being specifically designed for object localization and having access to only a fraction of the localization labels used by that method.}
    \label{tab:rvos}
     \vspace{-10pt}
\end{table*}

As SAM2 cannot always accurately segment the challenging and often ambiguous entities featured in Ref-VPS even with a large number of clicks, we manually refine the masks in frames where it fails (Figure~\ref{fig:annotation_process}, bottom). We additionally label ambiguous regions, such as the stem of the dandelion, as Ignore (shown in yellow). Visualizations of Ref-VPS annotations, together with additional statistics, are included in the supplementary. For evaluation, we follow~\citet{tokmakov2023breaking} and only report region similarity $\mathcal{J}$~\cite{everingham2010pascal} as contour accuracy $\mathcal{F}$~\cite{perazzi2016benchmark} is often not well defined for dynamic entities like smoke or light. Pixels inside the Ignore regions are not included in the metric calculation.

\section{Experiments}
\label{sec:experiment}
\smallsec{Datasets and evaluation.}
We evaluate our method on six benchmarks in total. Ref-YTB~\citep{seo2020urvos}, Ref-DAVIS~\citep{khoreva2019video}, and MeViS~\citep{ding2023mevis} are standard RVOS benchmarks, with MeViS focusing on challenging, motion‑guided referring expressions. For evaluating generalization to rare objects and `Stuff' categories, we use BURST~\citep{athar2023burst} and VSPW~\citep{miao2021vspw} datasets, respectively. 
Finally, we evaluate \methodname and the strongest baselines on our newly introduced Ref-VPS (detailed in Section~\ref{sec:dataset}). All the datasets, except Ref-YTB and MeViS, are only used for evaluation in a \textit{zero-shot} manner.

For Ref-YTB~\citep{seo2020urvos}, Ref-DAVIS~\citep{khoreva2019video}, and MeViS~\citep{ding2023mevis}, we use the standard evaluation metrics - Region Similarity ($\mathcal{J}$)~\cite{everingham2010pascal}, Contour accuracy ($\mathcal{F}$)~\cite{perazzi2016benchmark} and their mean ($\mathcal{J\&F}$)~\cite{davis2017}. For all other evaluations, we use the $\mathcal{J}$ metric. The evaluations on Ref-YTB and MeViS are done on the official servers, and we use the official metric implementation of Ref-DAVIS for all the other benchmarks.

\smallsec{Implementation details.} 
Our approach builds upon two state-of-the-art text-to-video diffusion architectures: ModelScope~\citep{wang2023modelscope} and Wan~\citep{wan2025wan}. Additional video diffusion backbones are evaluated in the supplementary. ModelScope comprises 1.4 billion parameters and extends Stable Diffusion~\citep{blattmann2023stable} with temporal modules. We adopt a two-stage training protocol following~\citet{zhu2024exploring}: in Stage I, we fine-tune only the spatial weights on Ref‐COCO image-text pairs\citep{yu2016modelingcontextreferringexpressions} for one epoch; in Stage II, we fine-tune all network weights for 40 epochs using Ref‐YTB video–text examples~\citep{seo2020urvos} supplemented with 12K Ref‐COCO images converted into pseudo‐videos following \citet{wu2022language}. In contrast, Wan employs a 14 billion parameter diffusion transformer that jointly models spatial and temporal information, without dedicated temporal modules~\citep{peebles2023scalable}. Accordingly, we train this variant in a single stage on the combined Ref‐COCO and Ref‐YTB datasets for 80k iterations. Throughout training, the text encoder and VAE remain frozen. Unless otherwise stated, all models are trained and evaluated at a resolution of $512\times512$.

More details about the datasets, evaluation setup on BURST and VSPW, implementation, as well as ablations of our training strategy and runtime analysis are included in the supplementary.

\subsection{Referring video object segmentation results}
\label{sec:experiment_refvos}
In this section, we compare \methodname to the state of the art on the standard RVOS benchmarks. We report results on the validation set of Ref-DAVIS~\citep{khoreva2019video} and the test set of Ref-YTB~\citep{seo2020urvos} in Table~\ref{tab:rvos}. The basic variant of our method, which is based on the ModelScope video diffusion model (denoted as `REM (MS-1.4B)' in the table), outperforms the state of the art in terms of $\mathcal{J}$ on Ref-DAVIS and is only second to UNINEXT~\citep{yan2023universal} on Ref-YTB. Note that this approach is specifically designed for object segmentation and trained on more than 10 datasets with localization annotations. In contrast, \methodname adopts an architecture of a video generation model and is only fine-tuned on one image- and one video-segmentation dataset. Despite this, our method is competitive with UNINEXT on standard RVOS benchmarks, and, as we will show next, outperforms it by up to 21 points out-of-domain in terms of $\mathcal{J}$. 

Another notable observation is that this variant of \methodname outperforms VD-IT~\citep{zhu2024exploring}, which is built on top of the same MS-1.4B backbone, on both datasets. This result demonstrates the effectiveness of our approach to preserving the generative model's architecture, which will become even more evident in out-of-domain evaluation. Using the recent, large-scale Wan T2V diffusion model~\cite{wan2025wan} (denoted as `REM (Wan-14B)') further improves performance, achieving state-of-the-art results on both benchmarks.

We also benchmark our \methodname against the top entrants on the MeViS leaderboard~\citep{ding2023mevis}, a dataset specifically designed for segmenting objects based on referring expressions describing their motion. As shown in Table~\ref{tab:mevis}, our base variant already outperforms the strongest baseline, GLUS~\citep{lin2025glusgloballocalreasoningunified}, which combines a pre-trained multi-modal LLM with the SAM2 foundational video segmentation model~\citep{ravi2024sam}, by 2.2 points in $\mathcal{J} \& \mathcal{F}$. Our Wan-14B variant further improves the performance to $60.3$, setting the new state-of-the-art for this benchmark. These results validate our method's ability to reuse the motion–language priors encoded in video diffusion models by preserving their full generative architecture.

\begin{figure*}[t]
    \centering
    \includegraphics[width= \textwidth]{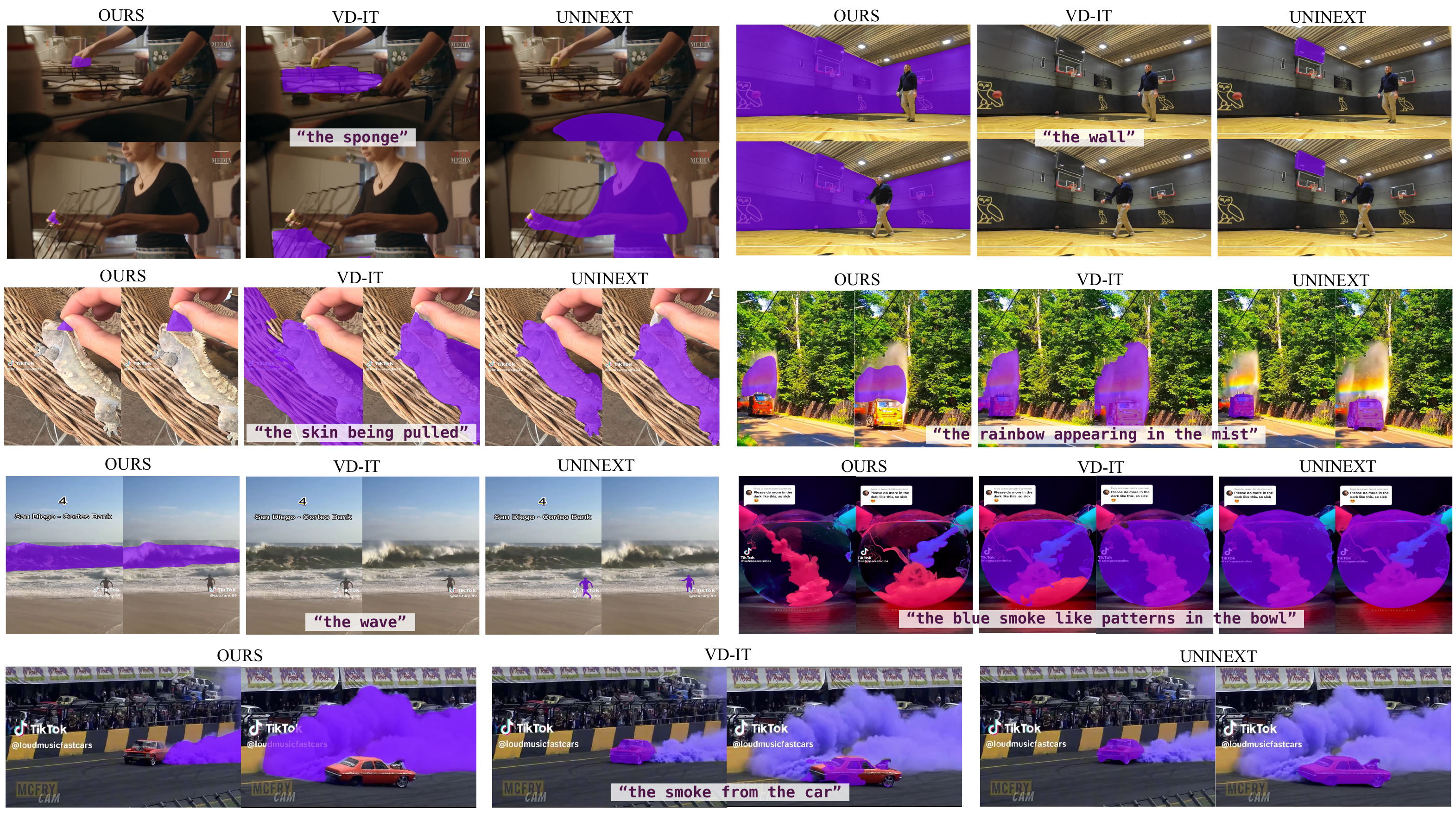}
    \vspace{-17pt}
    \caption{Qualitative results of \methodname (MS-1.4B) and state-of-the-art baselines on BURST (top row left), VSPW (top row right), and Ref-VPS (bottom three rows) benchmarks. Our method demonstrates both superior coverage of rare, dynamic concepts, such as smoke or waves, and higher segmentation precision (\eg, only capturing the skin on the lizard). Video comparisons are available \href{https://refereverything.github.io/\#dataset}{here}.}
    \label{fig:rvos}
    \vspace{-10pt}
\end{figure*}

\subsection{Out-of-domain generalization}
\label{sec:experiment_refproc}
\smallsec{Rare objects and `Stuff'.} We begin by performing a generalization study on the existing open-world tracking BURST dataset~\citep{athar2023burst} as well as on the `Stuff' categories~\citep{caesar2018coco} from VSPW~\citep{miao2021vspw} in Table~\ref{tab:stuff}. BURST is a large-scale, open-world video object segmentation benchmark with diverse scenes and rare objects, whereas VSPW tests the ability to generalize to non-object categories. We report \textit{zero-shot} evaluation results on the validation set of VSPW and combined validation and test sets of BURST and compare to the top-performing methods in Table~\ref{tab:rvos}.

We observe that on both datasets our method outperforms the baselines by significant margins. The improvements are especially noticeable on BURST, demonstrating that REM successfully preserves the object representation learned in generative pre-training. In contrast, VD-IT loses this generalization capacity. We analyze per-category performance in Section B.8 in supp., showing that \methodname is especially effective for the most challenging objects. On the `Stuff' categories, all the methods do relatively poorly, reflecting the challenge of generalizing to more amorphous `Stuff'. Here, VD-IT maintains a lead over entirely object-centric UNINEXT, but \methodname still outperforms both.

\begin{table}[t]
    \centering
    \resizebox{0.8\linewidth}{!}{
    \begin{tabular}{l|ccc}
         \multirow{2}{*}{Method} & \multicolumn{3}{c}{MeViS}   \\ 
         \cline{2-4} 
         & $\mathcal{J} \& \mathcal{F}$  & $\mathcal{J}$ & $\mathcal{F}$   \\
        \hline
         Referformer~\citep{wu2022language} & 31.0 & 29.8 & 32.2 \\
         VISA-13B~\citep{yan2024visareasoningvideoobject}  & 44.5 & 41.8 & 47.1 \\
         DsHmp~\citep{he2024decouplingstatichierarchicalmotion}  & 46.4 & 43.0 & 49.8 \\
         GLUS~\citep{lin2025glusgloballocalreasoningunified} & 51.3 & 48.5 & 54.2 \\
         \hline
         \methodname (MS-1.4B)   & 53.5 & 50.8  & 56.3 \\
         \methodname (Wan-14B)  & \bf 60.3  & \bf 57.2 & \bf 63.4 \\
         
    \end{tabular}
    }
    \caption{Comparison to the state of the art on the MeViS benchmark. REM outperforms approaches specifically designed for this dataset without any modifications.}
    \label{tab:mevis}
     \vspace{-10pt}
\end{table}

\begin{table}[t]
    \centering
\resizebox{\linewidth}{!}{
\begin{tabular}{l|ccc|cc}
\multirow{2}{*}{Benchmark} & \multirow{2}{*}{MUTR} & \multirow{2}{*}{UNINEXT} & \multirow{2}{*}{VD-IT} & \multicolumn{2}{c}{\methodname (Ours)} \\
\cline{5-6}
 &  &  &  & MS-1.4B & Wan-14B \\ \hline
VSPW & 10.5 & 10.1 & 12.7 & 15.2  & \bf 18.5\\
BURST & 27.2 & 30.2 & 29.0 & 37.5 & \bf 40.9
    \end{tabular}
    }
    \caption{Comparison against the state of the art on the `Stuff' categories in VSPW and on the open-world object-tracking BURST benchmark shows that \methodname achieves substantially stronger generalization. In particular, it outperforms VD-IT, which shares the same underlying diffusion backbone.}
    \label{tab:stuff}
    \vspace{-5pt}
\end{table}

\begin{table}[t]
    \centering
    \resizebox{\linewidth}{!}{

    \begin{tabular}{cccc|cc}
    \multirow{2}{*}{MUTR} & \multirow{2}{*}{UNINEXT} & \multirow{2}{*}{VD-IT} & \multirow{2}{*}{GLUS} & \multicolumn{2}{c}{\methodname (Ours)} \\
    \cline{5-6}
          &  &  &  & MS-1.4B & Wan-14B \\ \hline
          25.4 & 28.7 & 37.9 & 34.6 & 49.0 & \bf 50.0  \\
    \end{tabular}
    
    }
    \caption{Comparison to the state of the art on the new Ref-VPS benchmark. \methodname shows much stronger zero-shot generalization to challenging, dynamic concepts in this dataset compared to the baselines by effectively capitalizing on Internet-scale pre-training.}
    \label{tab:res-our-data}
    \vspace{-10pt}
\end{table}

\smallsec{Dynamic processes.} We compare \methodname to the top-performing RVS baselines on our Ref-VPS benchmark in Table~\ref{tab:res-our-data}. As before, all the evaluations are zero-shot. 
Our approach outperforms all baselines by up to 12.1 points in Region Similarity (31.9\% relative improvement), and notably surpasses the top RVOS method, UNINEXT, by 21.3 points (74.2\% relative improvement). While generative pre-training enhances VD-IT’s generalization ability compared to UNINEXT, it struggles to preserve its representations as effectively as REM. The recent GLUS approach, which was designed for MeViS, also fails to generalize to Ref-VPS, highlighting the complementarity of the two benchmarks.

A qualitative comparison of \methodname (MS-1.4B) with VD-IT and UNINEXT is provided in Figure~\ref{fig:rvos}. In the first row, our approach is able to track the sponge (which was never seen in training) in a challenging sequence from BURST, whereas other methods focus on foreground objects. In the second sequence from VSPW \methodname successfully generalizes to the non-object `wall' category, whereas UNINEXT focuses on a nearby \textit{object} and VD-IT fails entirely. The following examples from Ref-VPS illustrate that both baselines exhibit object-centric bias, as in the examples with the lizard skin in row 3 and blue smoke in row 6. While VD-IT shows better generalization, it often latches on the dominant region (rows 4 and 7 in Figure~\ref{fig:rvos}). In contrast, \methodname demonstrates both good coverage of rare concepts and high precision with respect to the language prompt. See more examples of highly dynamic sequences in the supplementary.

\subsection{Ablation analysis}
\label{sec:abl}
We now assess how effectively our \methodname transfers generative representations to RVS. We report results on the in-distribution Ref-YTB dataset and our Ref-VPS benchmark, using the MS-1.4B variant of our model. We further ablate the effect of the generative pre-training strategy in the supplementary. For efficiency, we train our model on a reduced subset of $\sim$12K image and video samples for these experiments, which accounts for the slightly lower performance compared to the full-data results reported above.

\smallsec{Is supervising in the latent space important?} We evaluate the effectiveness of our key design decision to supervise mask prediction in the latent space of the frozen VAE decoder in Table~\ref{tab:ablation-decode}. Instead of the latent space, we supervised the masks in the raw pixels space (RGB) by propagating the gradients through the VAE decoder (rows two and three in the table). This increases memory requirements during training, forcing a lower $256\times256$ input resolution to fit on an A100 GPU. The results show that latent-space supervision is crucial for maintaining generalization, as evidenced by the performance drop of the RGB-based variants on Ref-VPS, even when the VAE decoder is additionally finetuned.

\smallsec{Is it better to train a mask decoder from scratch?} Several prior works have proposed to use a de-noising network as a feature extractor and learn a mask decoder head from scratch~\citep{vpd23,zhu2024exploring}. Our approach uses a pre-trained VAE instead, which is not only a highly effective, general-purpose image encoder, but is also used in the de-noising network pre-training. We now ablate this design choice by replacing the VAE with a dedicated mask prediction module. We ablate both a CNN mask decoder from~\citep{vpd23} and an MLP decoder from~\cite{xie2021segformer} and train them jointly with the rest of the model at the full resolution (last 2 rows in Table~\ref{tab:ablation-decode}). Removing the pre-trained VAE has a moderately negative effect on performance on Ref-YTB, and, notably, destroys the model's ability to generalize to the out-of-distribution Ref-VPS (10.6 and 6.9 points drop for CNN and MLP, respectively). These results underscore the main takeaway of our analysis - preserving the entirety of the generative model's architecture is key for maximizing generalization in RVS.

\begin{table}[t]
    \centering
    \resizebox{\linewidth}{!}{
    \begin{tabular}{l|c|c|c|c}

        \multirow{2}{*}{Supervision} & \multirow{2}{*}{Resolution} & \multirow{2}{*}{Decoder} & Ref-YTB  & Ref-VPS \\
        & & & ($\mathcal{J}\&\mathcal{F}$) & $\mathcal{J}$ \\
        \hline
        Latent  & 256$\times$256 & VAE (Frozen) & \textbf{61.4} & \textbf{35.8}\\
        RGB & 256$\times$256 & VAE (Frozen) & 58.4 & 31.6 \\
        RGB & 256$\times$256 & VAE (Fine-tuned) & 60.4 & 32.4 \\
        \hline
        Latent & 512$\times$512 & VAE (Frozen) & \bf 63.5 & \bf 40.0 \\
        RGB & 512$\times$512 & CNN  & 59.6 & 29.4 \\
        RGB & 512$\times$512 & MLP & 59.3 & 33.1 
        
    \end{tabular}
    }
    \caption{Analysis of the effect of the fine-tuning strategy on Ref-YTB and Ref-VPS. The key to the success of \methodname is in preserving the entirety of the generative model’s architecture. }
    \label{tab:ablation-decode}
    \vspace{-12pt}
\end{table}

\section{Discussion}
We introduced \methodname, a framework that leverages Internet-scale video-language representations learned by diffusion models to enable open-world referring video segmentation. By preserving the full generative model architecture, \methodname preserves the universal visual-language mapping, allowing it to generalize beyond object-centric segmentation to dynamic, non-object concepts. Our experiments show strong generalization on standard datasets and our new Ref-VPS benchmark for dynamic video processes. Despite limited training on object masks, \methodname outperforms prior methods by up to 12 points in region similarity out-of-domain. Notably, we also show that advances in video diffusion models directly improve video segmentation performance.

\section*{Acknowledgement}

This project is supported in part by Toyota Research Institute, NSF Grant 2106825, and NIFA Award 2020-67021-32799.

{
    \small
    \bibliographystyle{ieeenat_fullname}
    \bibliography{arxiv_conference}
}

\clearpage
\newpage
\appendix
\setcounter{figure}{0}
\setcounter{table}{0}
\setcounter{equation}{0}
\renewcommand{\thefigure}{\Alph{figure}}
\renewcommand{\thetable}{\Alph{table}}
\renewcommand{\theequation}{\Alph{equation}}

In this supplementary material, we first include additional details for our Ref-VPS dataset in Section~\ref{suppsec:dataset}. 
Next, we offer a deeper quantitative analysis, covering VAE-based mask reconstruction, failure modes, and computational costs, \etc, in Section~\ref{suppsec:quantitative}. Section~\ref{suppsec:qualitative} presents additional qualitative evaluations, including visualizations on typical failure cases, challenging fight scenes, and ambiguous or overlapping scenarios. Finally, in Section~\ref{suppsec:implementation}, we report all the implementation details.

\section{Ref-VPS Dataset Details}
\label{suppsec:dataset}

\subsection{Dataset collection pipeline and statistics}
During our dataset collection, we first established a non-exhaustive taxonomy of five broad and possibly overlapping concepts. This taxonomy was designed to encompass key modes of dynamic change while offering a structured framework for the task. The concepts and their definitions are as follows:
\begin{itemize}
    \item \textbf{Temporal Object Changes:} Phenomena where an object’s state or shape evolves over time (\eg, object deformation, melting)
    \item \textbf{Motion Patterns:} Motion in amorphous or non-rigid regions (\eg, water ripples, flickering flames)
    \item \textbf{Dynamic Environmental Changes:} Environmental transformations affecting spatial regions over time (\eg, clouds moving across the sky, waves rising ) 
    \item \textbf{Interaction Sequences:} Events characterized by interactions between objects (\eg, bullet hitting glass, object collisions)
    \item \textbf{Pattern Evolution:} Progressive changes in patterns or textures (\eg, changing patterns of smoke dispersion, fluctuating light levels) 
\end{itemize}

\begin{figure}
    \centering
    \includegraphics[width = \linewidth]{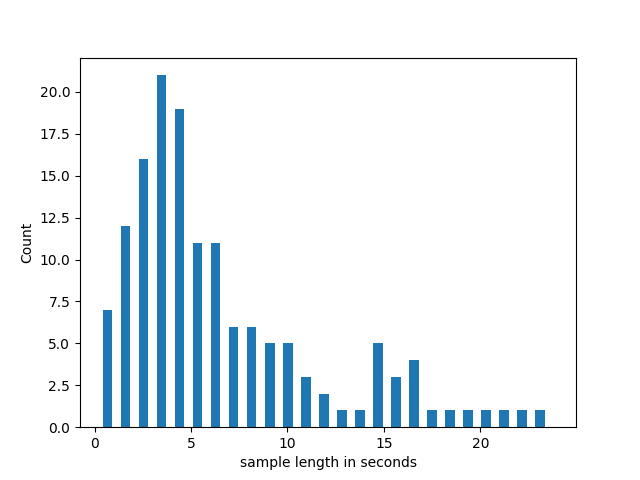}
    \vspace{-20 pt}
    \caption{Distribution of sample lengths in Ref-VPS. Most of our samples are between 2.5 to 5 seconds in length, but can go up to more than 20 seconds.}
    \label{fig:data_hist}
\end{figure}

Our final dataset comprises 145 video clips representing 39 distinct dynamic process concepts. We report a comprehensive list of key statistics in Table~\ref{tab:dataset}. Most of our samples are between 2.5 and 5 seconds in length, but can go up to more than 20 seconds. The distribution of our sample lengths is reported in Figure~\ref{fig:data_hist}.

\begin{table}[t]
     \centering
     \resizebox{0.6 \linewidth}{!}{
     \begin{tabular}{l|c} \hline
         Clips &  145 \\ 
         FPS & 24 \\
         Frames & 23,442\\ 
         Concepts & 39 \\ 
         Avg length (s) & 6.74 \\ 
         Annotation FPS & 6 \\ 
         Min-resolution & $712 \times 576$\\ 
         Max-resolution & $1024 \times 576$\\\hline
     \end{tabular}
     }

     \caption{Statistics of our Ref-VPS benchmark. Our dataset contains 145 video clips covering 39 concepts for dynamic processes.}
     \label{tab:dataset}
 \end{table}

\begin{figure*}[t]
    \centering
    \includegraphics[width = 0.95 \linewidth]{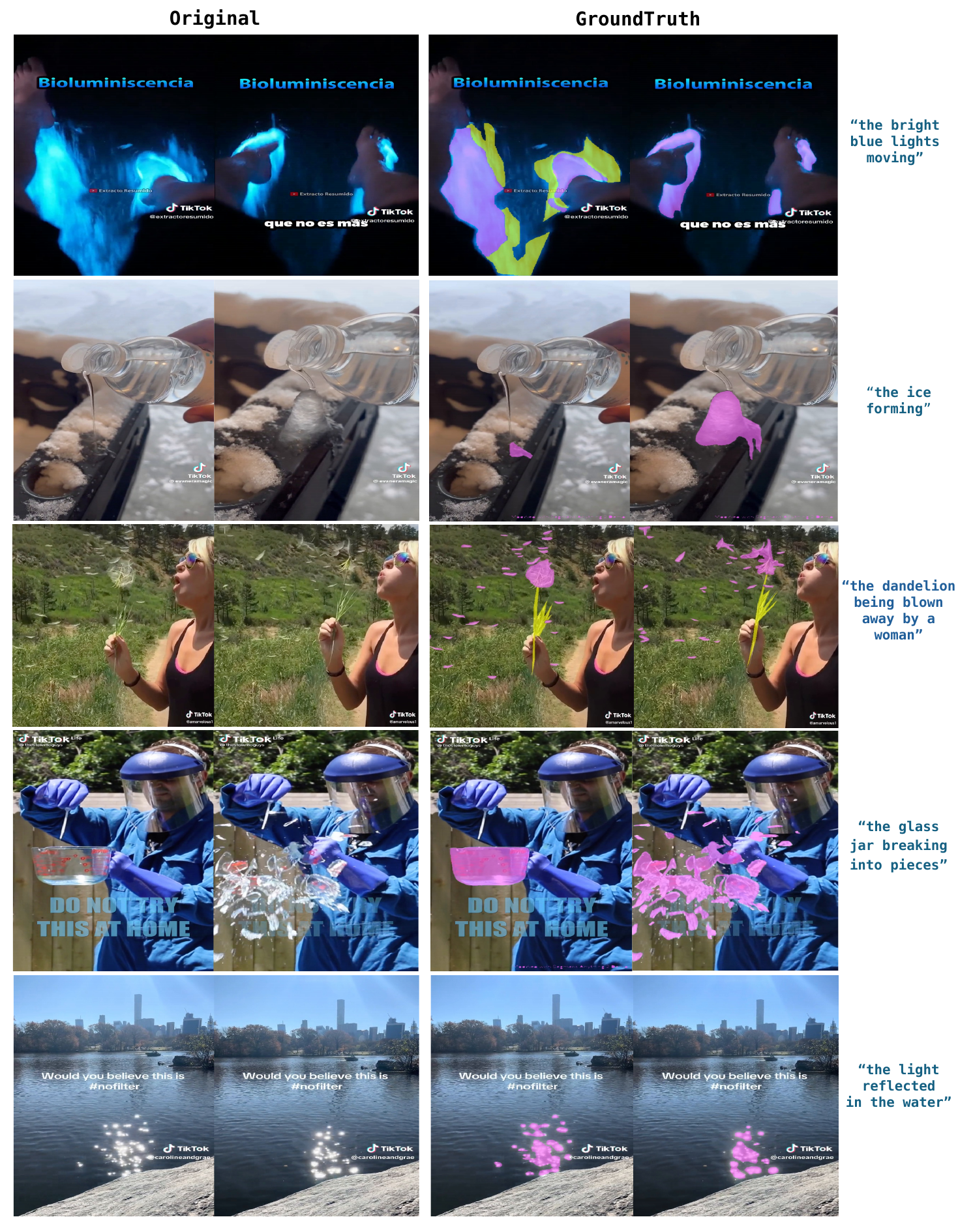}
    \vspace{-10 pt}
    \caption{Samples from our Ref-VPS dataset. Ground-truth masks are shown in pink, and the Ignore regions are shown in yellow. Pixels inside the Ignore regions are not included in the metric calculation.}
    \label{fig:refvpsannotation}
\end{figure*}
 
\subsection{Annotation visualizations}

Figure~\ref{fig:refvpsannotation} showcases examples of our Ref-VPS segmentation mask annotations. Our annotations capture the full extent of target objects, as seen with the icicle (second row) and the glass (fourth row). For more ambiguous cases, such as glowing water (first row) or a dandelion being blown (third row), only the confident regions are labeled, while uncertain areas are marked as Ignore (yellow). These \emph{Ignore Regions} are excluded from metric computation, ensuring that evaluations focus on reliable mask regions and are not penalized for inherently ambiguous boundaries.

\subsection{Mask annotation accuracy evaluation}

To assess the quality of our annotations, we compute inter-annotator agreement on the Ref-VPS dataset. Specifically, an independent annotator relabeled a subset of the dataset, covering all 39 dynamic concepts, using the same annotation protocol. Following the evaluation approach of \citet{benenson2019large}, we report an inter-annotator mean IoU (mIoU) of \textbf{87.1\%}, significantly higher than the $\sim$80\% agreement number reported for COCO~\citep{caesar2018coco}. This high agreement demonstrates the effectiveness of our annotation protocol, particularly the use of Ignore labels to handle ambiguity in subjective scenarios.

\section{Additional Quantitative Evaluations}
\label{suppsec:quantitative}

In this section, we provide additional quantitative evaluations for our proposed \methodname. Same as our ablation study in the main paper, we conduct these experiments using the MS-1.4B version, unless staged otherwise. 

\subsection{Mask reconstruction accuracy analysis} 

In designing our REM model, we repurpose a pre-trained VAE as the mask decoder, based on the intuition that large-scale pre-training enables the VAE to effectively reconstruct masks as images. To validate this assumption, we quantitatively evaluate the VAE's reconstruction performance on binary mask images, following the methodology of Marigold~\citep{ke2024repurposing}. Specifically, we assess reconstruction accuracy on 3,471 binary masks from the Ref-YTB training set (one per video). The VAE achieves a mean absolute error (MAE) of \textbf{0.0144} for mask reconstruction. In comparison, reconstructing the corresponding RGB frames yields a higher MAE of \emph{0.1236}, reflecting the greater difficulty of the RGB task. Furthermore, the VAE attains a mask reconstruction mIoU of \textbf{99.33\%} between the input and output masks. These results support our approach of converting masks into 3-channel inputs for compatibility with pre-trained auto-encoders, effectively mitigating concerns about domain mismatch.

\subsection{Failure mode analysis}

\begin{figure}
    \centering
    \includegraphics[width=\linewidth]{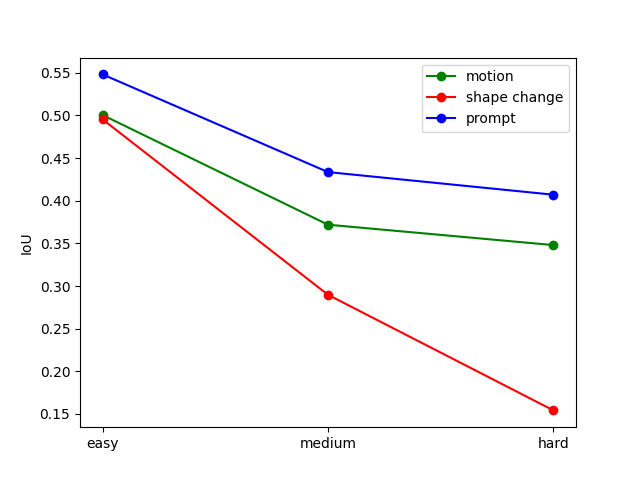}
    \vspace{-30 pt}
    \caption{Quantitative evaluation of our REM (MS-1.4B) failure modes on Ref-VPS: Significant shape change is the primary failure mode, with motion and prompt complexity having secondary impacts.}
    \label{fig:fail_quant}
\end{figure}

We conducted a quantitative evaluation of our REM failure modes on Ref-VPS in Figure~\ref{fig:fail_quant}. We measure performance degradation across motion and shape changes (following \citet{dave2020tao}), and prompt complexity (sentence length). Our analysis reveals that significant shape change is the primary failure mode, with motion and prompt complexity having secondary impacts. These results further illustrate the challenge of segmenting dynamic concepts in videos.

\subsection{Computational cost}

We report the inference speed and memory consumption of REM (MS-1.4B) alongside key baselines in Table~\ref{tab:comp_infer}, using their official public implementations. All measurements are conducted on 32-frame clips from Ref-DAVIS using a single NVIDIA A100 GPU, with averages computed over 80 runs. As shown, the inference costs of REM align with those of other state-of-the-art approaches.

For training, \methodname requires 174 hours on four A100 GPUs. Since most baselines do not disclose training costs, we estimate them under equivalent hardware and conditions (excluding I/O time), and report the results in Table~\ref{tab:comp_train}, including per-GPU memory consumption. Our training efficiency is on par with prior works. Notably, UNINEXT, the current state-of-the-art in RVOS, requires approximately 6.3 times longer to train than REM due to its reliance on over ten supervised datasets to achieve strong object segmentation performance. In contrast, REM leverages Internet-scale pre-training to attain comparable performance on in-domain benchmarks and significantly outperforms UNINEXT in out-of-distribution scenarios, all while incurring a fraction of the training cost.

\begin{table}[t]
\centering
    \resizebox{0.9 \linewidth}{!}{
    \begin{tabular}{l|cc}
        Method &  Memory (GB) & Speed (FPS)\\ \hline
        MUTR & 34.1 & 13.6 \\
        UNINEXT & 9.7 & 3.3  \\
        VD-IT & 72.8 & 7.1  \\ 
        \hline 
        \methodname (MS-1.4B) & 41.8 & 7.1  \\
    \end{tabular}
    }
    \caption{Inference costs of REM and top RVS methods on Ref-DAVIS. Both the memory requirements and the runtime of \methodname are on par with other models in the literature.
    }
    \label{tab:comp_infer}
\end{table}
\begin{table}
    \resizebox{\linewidth}{!}{
    \begin{tabular}{l|cc}
        Method &  Memory (GB) & Total Runtime (hr) \\ \hline
        MUTR & 30.4 &  134 \\
        UNINEXT & 30.2 & 1906  \\
        VD-IT & 68.5 &  260 \\ 
        \hline 
        \methodname (MS-1.4B) & 61.8 & 174 \\
    \end{tabular}
    }
    \caption{Training costs of REM and top RVS methods. Our costs are on par with prior work and are notably significantly lower compared to UNINEXT, the state-of-the-art RVOS approach. 
    }
    \label{tab:comp_train}
\end{table}

\subsection{Ablation of the training strategy}

Our model adopts a two-stage training strategy (detailed in Section~\ref{suppsec:implementation}), where we first pre-train on image-only data to learn spatial representations, followed by joint fine-tuning on mixed image and video data. In this section, we compare the two-stage approach to a single-stage alternative and analyze the impact of varying the amount of image data used in the second stage.

\smallsec{Benefits of two-stage training.} 
As reported in Table~\ref{tab:ablation-stage}, the two-stage training strategy yields superior performance on both the standard RVS benchmark and the out-of-distribution Ref-VPS dataset. This improvement stems from allowing the model to first acquire strong spatial priors from image-only data before incorporating the more complex temporal dynamics of videos. Additionally, initializing with well-trained spatial weights enhances training stability and convergence.

\smallsec{Impact of image data used in the second stage}. 
In our default two-stage setup, we use an equal amount of image and video data during the second stage. To investigate the effect of image data volume, we consider two variants: one with twice as many images and one with no image data. As shown in Table~\ref{tab:ablation-pseudo}, using no image data significantly degrades generalization on Ref-VPS, underscoring the importance of image supervision. Conversely, doubling the image data leads to performance degradation on both benchmarks, suggesting that excessive reliance on static visual information can hinder the learning of spatiotemporal dynamics. These results highlight the importance of a balanced integration of image and video data for effective training.

\begin{table}[t]
    \centering
    \resizebox{\linewidth}{!}{
    \begin{tabular}{l|c|c|c}

        \multirow{2}{*}{Training stages} & \multirow{2}{*}{Training strategy} & Ref-YTB  & Ref-VPS \\
        & & ($\mathcal{J}\&\mathcal{F}$) & $\mathcal{J}$ \\
        \hline
        Two-stages (Ours)  & Images $\rightarrow$ Images \& Videos   & \textbf{68.4} & \textbf{49.0}\\
        Single-stage & Images \& Videos & 66.3 & 46.33 
    \end{tabular}
    }
    \caption{Comparison between our default two-stage training and single-stage training strategy. The two-stage training strategy allows the model to first learn strong spatial representations from image-only data before incorporating the more complex temporal dynamics present in videos. Therefore, it achieves better results for both the standard RVS benchmark and our out-of-distribution Ref-VPS dataset. 
    }
    \label{tab:ablation-stage}
\end{table}

\begin{table}[t]
    \centering
    \begin{tabular}{l|c|c}

        \multirow{2}{*}{Images:Videos } & Ref-YTB  & Ref-VPS \\
        & ($\mathcal{J}\&\mathcal{F}$) & $\mathcal{J}$ \\

        \hline
         2:1    & 67.0 & 48.0\\
         \textbf{1:1 (Ours)} & 68.4 & \textbf{49.0} \\
         No images &  \textbf{68.7} & 40.7

    \end{tabular}
    \caption{Impact of the image data volume used in the second stage. Incorporating image data in the second stage significantly enhances the model’s generalization on the Ref-VPS dataset compared to the version trained without images, while using more image data yields suboptimal results. Overall, a balanced integration of image and video data is key to the success of our approach.}
    \label{tab:ablation-pseudo}
\end{table}

\subsection{Effect of generative pre-training on RVS}

We focus on how generative pre-training affects the RVS performance in this section. We focus on comparing our MS-1.4B model variants with other pre-trained diffusion models of a similar parameter size. 

We begin by evaluating the effect of Image generation pre-training in Table~\ref{tab:ablation-backbone}. As a baseline, we first fine-tune Stable Diffusion 2.1 ~\citep{blattmann2023stable} (an image generation model, denoted as SD2.1) on individual frames (column 1 in the table). This variant has no temporal modeling capacity, but neither does UNINEXT~\citep{yan2023universal} - the state-of-the-art approach for RVOS. However, it strongly underperforms compared to our best video-based variants, not only on Ref-VPS but also on the object-centric Ref-YTB. This shows that while generative pre-training relies heavily on images, video data is crucial for learning effective representations for tracking.

Next, we evaluate two variants of the VideoCrafter model~\citep{chen2023videocrafter1,chen2024videocrafter2} (denoted as VC-1 and VC-2 in Table~\ref{tab:ablation-backbone}), both initialized from Stable Diffusion 2.1~\citep{blattmann2023stable} and trained on 600M images and 10-20M videos. VC-2 focuses on high-quality data curation, which has been shown to be important for representation learning in the past~\cite{radford2021learning, gadre2023datacomp}, and leads to substantial performance gains across both benchmarks.
Finally, the ModelScope~\citep{wang2023modelscope} approach is also initialized from Stable Diffusion, but trained on the larger LAION 2B and a similar amount of high-quality video data (last column in Table~\ref{tab:ablation-backbone}). It performs comparably to VC2 on Ref-YTB, while demonstrating the best zero-shot generalization to Ref-VPS among all the variants, making it our default representation. These results highlight that large-scale image pre-training, combined with generative video-language modeling, is important for generalization in RVS.

\begin{table}[t]
    \centering
    \resizebox{\linewidth}{!}{
    \begin{tabular}{l|cccc}
    Benchmark & SD2.1 & VC-1 & VC-2 & MS-1.4B\\ \hline 
    Ref-YTB ($\mathcal{J}\&\mathcal{F}$) & 60.2 & 57.5 &  64.9 & \bf 63.5  \\
    Ref-VPS ($\mathcal{J}$) & 29.8 & 28.0 & 36.8 &  \bf 40.0

    \end{tabular}
    }
    \vspace{-7pt}
    \caption{Analysis of the effects of generative pre-training on Ref-YTB and Ref-VPS. Both large-scale image pre-training as well as learning to model video-language interactions are important for robust RVS performance.}
    \label{tab:ablation-backbone}
    \vspace{-7pt}
\end{table}

\begin{table}[t]
    \centering
    \begin{tabular}{l|cc}
Noise Level & Ref-YTB ($\mathcal{J} \& \mathcal{F}$) & Ref-YTB ($\mathcal{J}$)\\ \hline
        200 & 59.2 & 36.2 \\
        50 & \bf 62.9 & 35.0\\ \hline
        0 (Ours) & \bf 62.9 & \bf 40.5\\
    \end{tabular}
    \caption{Ablation study on the choice of the noisy timestep. The best performance is achieved with minimal noise ($t = 0$), validating our design.}
    \label{tab:ablation-noise}
\end{table}

\begin{table}[t]
\centering
    \centering
    \resizebox{\linewidth}{!}{
\begin{tabular}{l|cc|cc}
    \multirow{2}{*}{Method} & \multicolumn{2}{c|}{Ref-VPS} & \multicolumn{2}{c}{Ref-DAVIS} \\ \cline{2-5}
    & $\mathcal{J}$ & Temp. Con.  & $\mathcal{J}$ & Temp. Con. \\ \hline
    MUTR &  24.1 &  2.9   &   64.8   &  3.4\\
    UNINEXT & 26.3 & 5.2 & 68.2 & 5.2\\
    VD-IT & 35.3 & 4.7 & 66.2 & 3.1\\
    \hline
    \methodname (MS-1.4B) & 49.0 & 2.8 & 69.9 & 2.1 \\
    
\end{tabular}
}
    \caption{Temporal Consistency comparison to the state of the art on Ref-VPS and Ref-DAVIS. Our approach demonstrates the best temporal consistency on both object-centric and non-object-centric datasets.}
    \label{tab:temp_cons}
\end{table}

\begin{table*}[t]
    \centering
    \resizebox{\linewidth}{!}{
    \begin{tabular}{l|l|ccc}
         \multirow{2}{*}{Method} & \multirow{2}{*}{Mask annotation} & \multicolumn{3}{c}{MeViS}   \\ 
         \cline{3-5} 
         & & $\mathcal{J} \& \mathcal{F}$  & $\mathcal{J}$ & $\mathcal{F}$   \\
        \hline
         Referformer~\citep{wu2022language} & MeViS & 31.0 & 29.8 & 32.2 \\
         VISA-13B~\citep{yan2024visareasoningvideoobject}  & Ref-COCO/+/g, Ref-YTB, MeViS, Ref-DAVIS, ReVOS, LVVIS, Refclef, ADE20k & 44.5 & 41.8 & 47.1 \\
         DsHmp~\citep{he2024decouplingstatichierarchicalmotion}  & MeViS & 46.4 & 43.0 & 49.8 \\
         GLUS~\citep{lin2025glusgloballocalreasoningunified} & Ref-YTB, MeViS, Ref-DAVIS, ReVOS, LVVIS& 51.3 & 48.5 & 54.2 \\
         \hline
         \methodname (Wan-14B)  & MeViS & 57.6  & 54.3 & 60.9 \\
         \methodname (Wan-14B)  & Ref-COCO/+/g, Ref-YTB, MeViS & \bf 60.3  & \bf 57.2 & \bf 63.4 \\
         
    \end{tabular}
    }
    \caption{Comparison to the state of the art on the MeViS benchmark with a comprehensive list of mask annotations used in training. Our REM (Wan-14B) reaches a new state of the art on MeViS while relying on orders of magnitude fewer pixel-level annotations than prior work.}
    \label{tab:mevis_details}
     \vspace{-10pt}
\end{table*}

\subsection{Ablation study on the noisy timestep}

In the main paper, we default to a noise timestep of $t = 0$, based on the observation that our task formulation focuses on direct mask latent prediction rather than denoising. This design choice eliminates the need for injecting noise into the latent space. To empirically validate this decision, we conduct a single-stage training experiment on the Ref-YTB dataset and report the results in Table~\ref{tab:ablation-noise}. The findings confirm our hypothesis: minimal noise (\ie, $t = 0$) consistently leads to the best performance, reinforcing the suitability of this choice for our predictive framework.

\subsection{Temporal consistency evaluation}

Evaluating temporal consistency in video segmentation remains a challenging task, as it is difficult to disentangle variations caused by model inconsistency from those arising due to genuine object deformations. For example, the temporal consistency metric initially introduced in the DAVIS dataset~\citep{davis2017} was applied only to videos with minimal object deformation and occlusion, and was eventually deprecated by the dataset authors due to its limited applicability.

To address these challenges, we adopt a simple yet effective temporal consistency metric that quantifies frame-to-frame stability. Specifically, we compute the average difference in Intersection-over-Union (IoU) between predicted masks and ground truth masks across consecutive frames. Formally, the metric is defined as:
\begin{equation}
    \text{Temp.~Con.} = \frac{1}{N}  \sum_{n=1}^N \left [\frac{1}{T_n} \sum_{t=1}^{T_n}(IoU_\text{diff}) \right ],
\end{equation}
where $N$ is the number of samples and $T_n$ is the number of frames in the $n^{th}$ sample, and 
\begin{equation}
    IoU_\text{diff} = IoU(Pred_{t+1}, GT_{t+1}) - IoU (Pred_t, GT_t) .
\end{equation}
Lower values indicate better temporal consistency. However, it is important to interpret this metric in conjunction with prediction accuracy, as trivially empty predictions would yield a perfect consistency score of zero without meaningful segmentation.

We report both region similarity and temporal consistency on Ref-VPS and Ref-DAVIS (both sampled at 24 fps) in Table~\ref{tab:temp_cons}. REM achieves the best temporal consistency on both object-centric and non-object-centric datasets. Interestingly, while MUTR also attains a strong consistency score on Ref-VPS, this is primarily due to its frequent output of empty masks, as reflected in its low region similarity. Conversely, UNINEXT, despite being the state-of-the-art on traditional RVOS benchmarks, shows the poorest temporal stability across both datasets.

\subsection{Concept coverage plot on BURST dataset}
\begin{figure}[t]
    \centering
    \includegraphics[width = \linewidth]{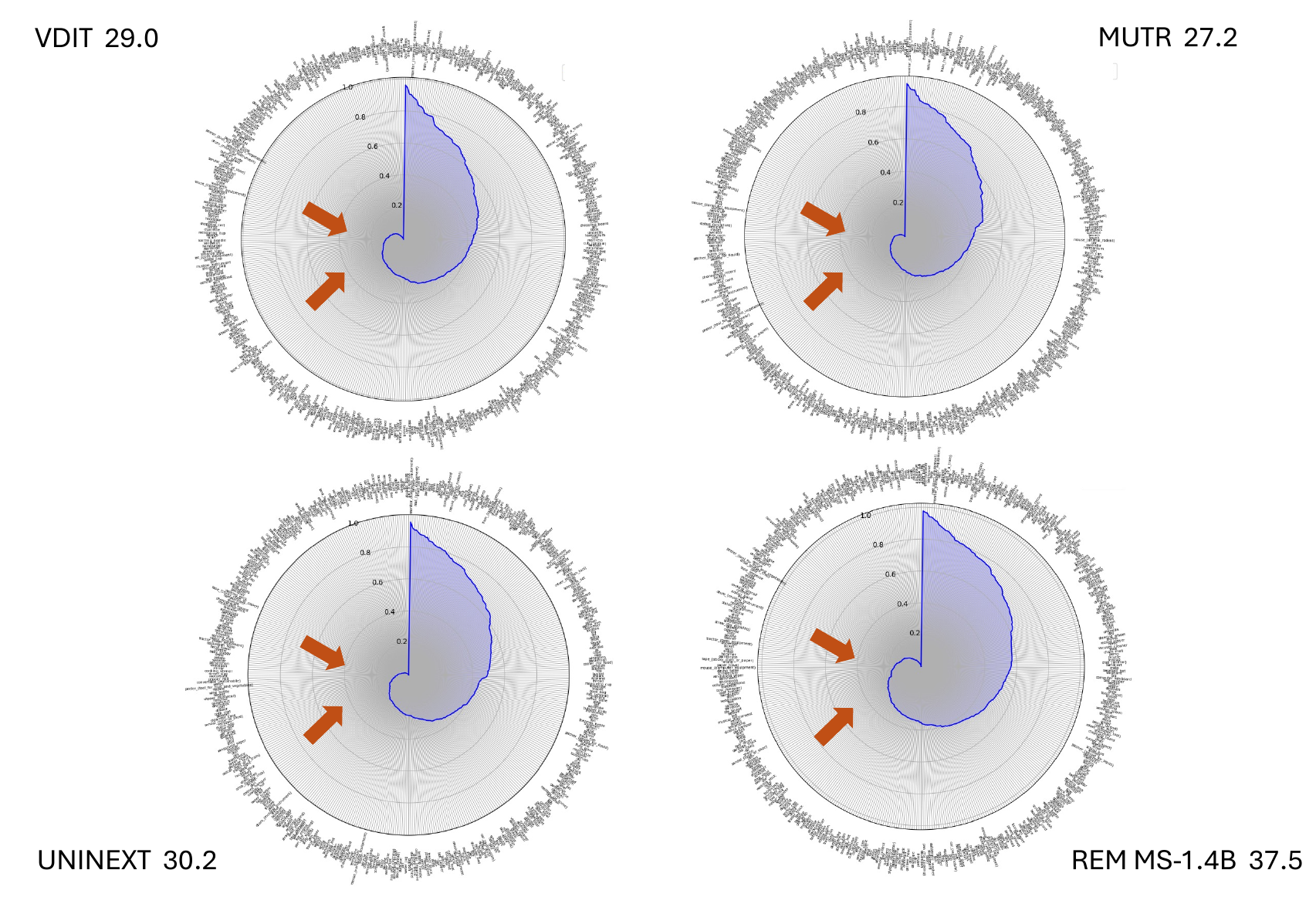}
    \caption{ Class-wise $\mathcal{J}$ scores (mIoU) across 454 object classes demonstrating concept coverage on BURST. As indicated by the arrows, REM is more robust on the most challenging categories compared to other methods.}
    \label{fig:coverage}
\end{figure}

Figure~\ref{fig:coverage} presents the concept coverage plots on the BURST~\citep{athar2023burst} dataset for VD-IT~\citep{zhu2024exploring}, MUTR~\citep{yan2024referred}, UNINEXT~\citep{yan2023universal}, and our method. As shown, \methodname consistently outperforms the baselines on the most challenging categories, further highlighting its strong generalization ability across a diverse range of visual concepts.

\subsection{Additional evaluation on MeViS}

\smallsec{Data efficiency.} We first provide the data source for all the baseline models we compared in Table~\ref{tab:mevis_details}. Our REM (Wan-14B) reaches a new state of the art on MeViS while relying on \emph{orders of magnitude fewer} pixel-level annotations than prior work. In particular, the strongest baseline, GLUS, couples a large-scale multimodal LLM with SAM2 masks and aggregates supervision from at least six video- and image-level datasets. In contrast, the full version of \methodname is fine-tuned on just three datasets, yet it improves the previous best $\mathcal{J} \& \mathcal{F}$ from 51.3\% to 60.3\%. These results underline that preserving the generative architecture of a diffusion model transfers rich visual–language priors so effectively that only a modest amount of task-specific data is needed to surpass far heavier-supervised baselines.

\smallsec{Single-dataset training.} To isolate the contribution of our training protocol, we additionally built a variant of our Wan model by finetuning on the MeViS training set alone. Performance decreases by only 2.7 in terms of $\mathcal{J} \& \mathcal{F}$, yet it still outperforms the strongest published baseline by 6.3 points. This resilience demonstrates that our mixed training strategy endows the model with robust spatial–temporal representations that generalize even when the downstream supervision is extremely limited. We include the training details for our Wan variant in Section~\ref{suppsec:implementation}.

\section{Additional Qualitative Evaluations}
\label{suppsec:qualitative}

\subsection{Failure cases visualizations}
\label{sppsec:fail}
\begin{figure}[t]
    \centering
    \includegraphics[width = \linewidth]{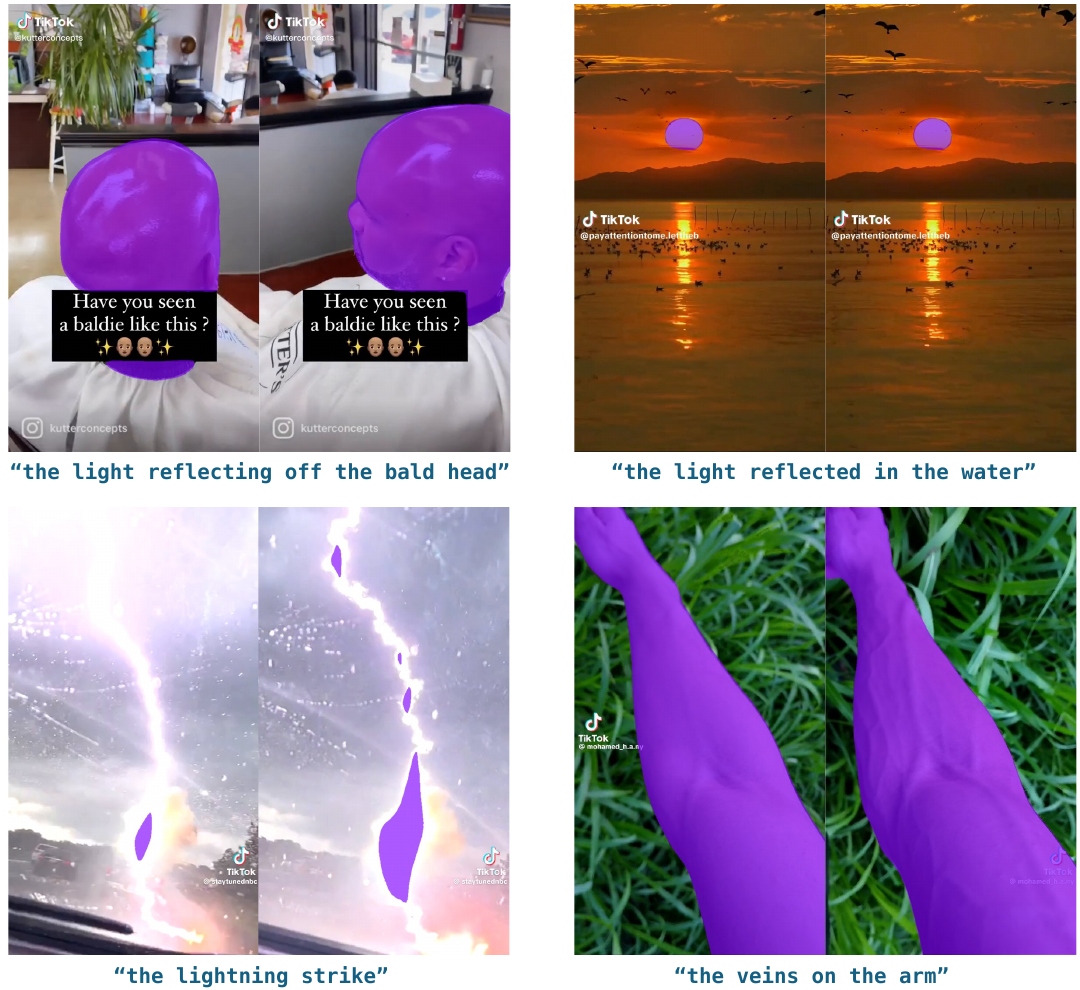}
    \vspace{-20pt}
    \caption{Failure cases of \methodname (MS-1.4B) on Ref-VPS. Our model still exhibits some object-centric bias and struggles with extremely dynamic entities such as lightning. }
    \label{fig:failure}
\end{figure}

A few representative failure cases of \methodname (MS-1.4B) on Ref-VPS are shown in Figure~\ref{fig:failure}. Our method suffers from object-centric bias in the most challenging scenarios (\eg, light reflection and veins) and struggles with extremely fast motion (\eg, the lightning strike).

\subsection{Evaluation on challenging fight scenes}
\label{sppsec:fight}

Fight sequences in movies and animated shows present a particularly challenging setting for referring video segmentation. These scenes are often characterized by severe and frequent occlusions, objects or characters exiting the frame, and rapid camera pose changes. Such factors cause drastic variations in appearance, demanding high temporal and semantic consistency to accurately track, re-identify, and segment the referred entities.

\begin{figure*}[t]
    \centering
    \includegraphics[width = 0.97 \linewidth]{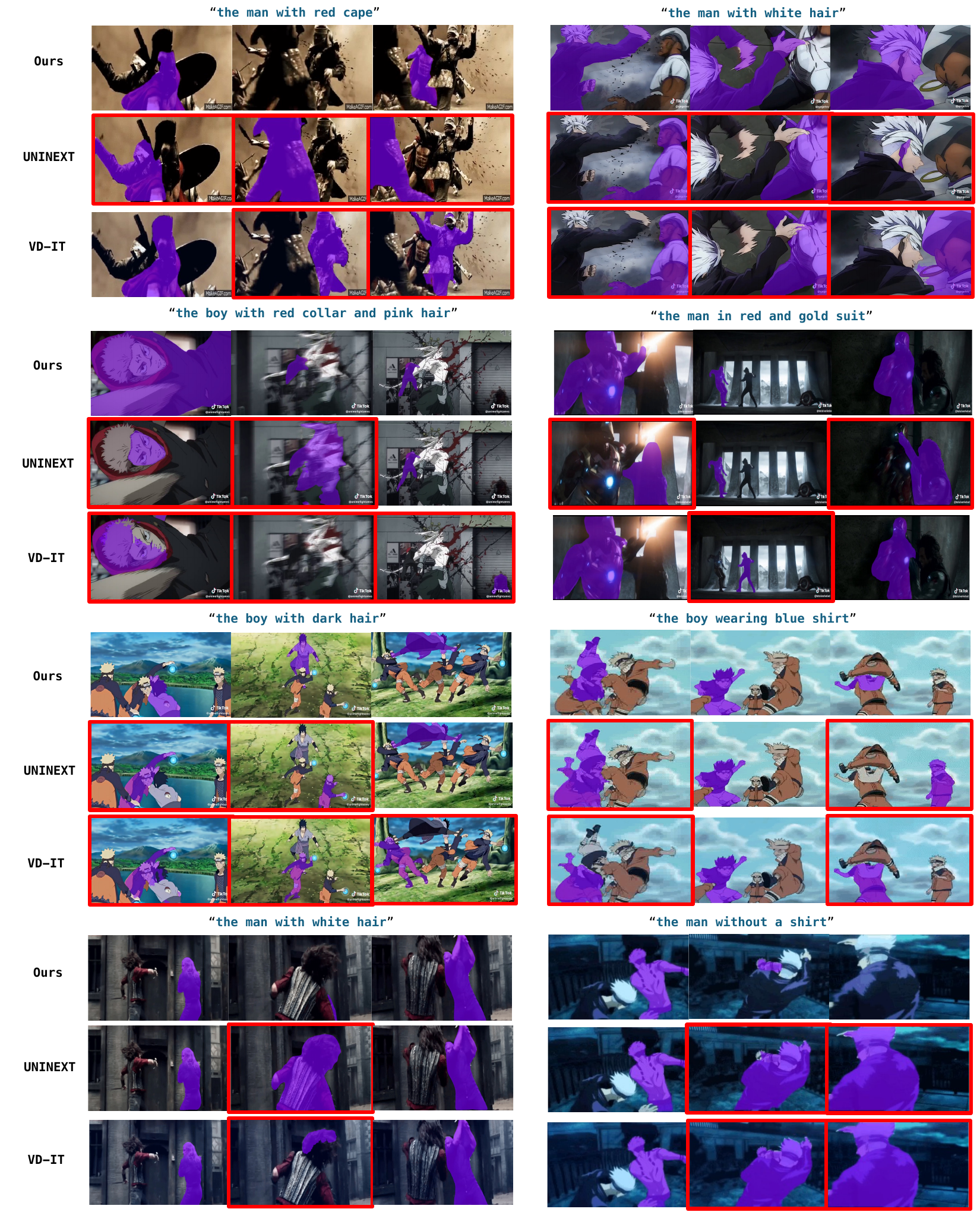}
    \vspace{-5 pt}
    \caption{ Qualitative comparison of REM (MS-1.4B) with state-of-the-art baselines on dynamic and challenging fight scenes. The incorrectly segmented frames are outlined in red. REM outperforms the other methods in handling frequent occlusions and POV changes. For a better illustration of the differences, please watch the full videos \href{https://refereverything.github.io/\#fight}{here}.}
    \label{fig:fightting}
\end{figure*}

Our \methodname excels in this domain of extremely challenging samples as illustrated in Figure \ref{fig:fightting}. In contrast, both UNINEXT and VD-IT exhibit clear failure cases when the referred entity undergoes large occlusions or momentarily disappears from view. Notably, despite utilizing a video diffusion backbone, VD-IT fails to fully exploit the temporal consistency learned during video diffusion pre-training, whereas REM maintains robust performance under these challenging conditions.

\begin{figure}[t]
    \centering
    \includegraphics[width = \linewidth]{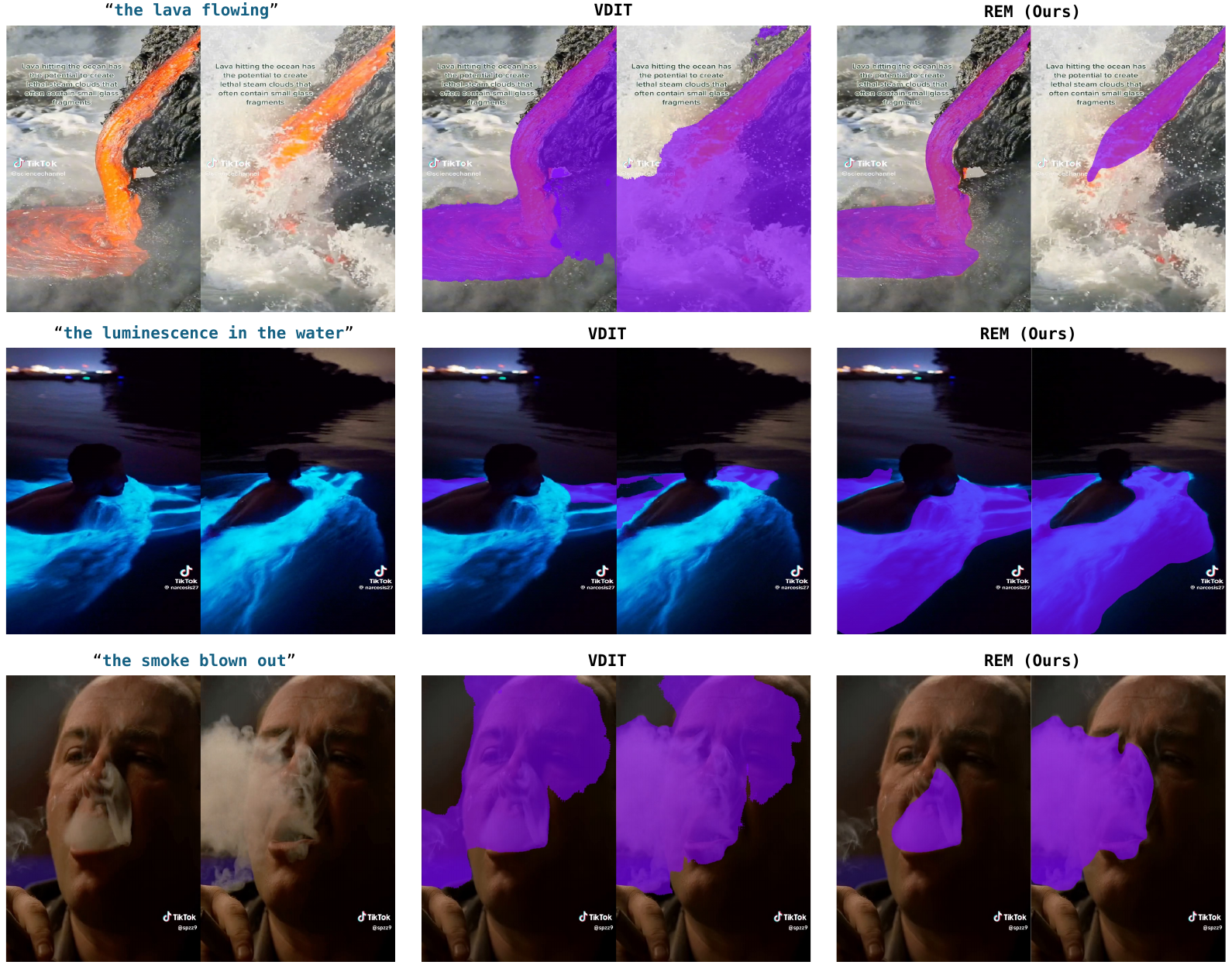}
    \caption{Comparison on ambiguous or overlapping scenarios in Ref-VPS between VD-IT and \methodname (MS-1.4B). While no single perfect prediction exists for these samples, our method is both more precise and more consistent.}
    \label{fig:ambiguity}
\end{figure}

\subsection{Comparisons on ambiguous or overlapping scenarios}

To assess how well our method handles visually ambiguous or overlapping scenarios, we present a qualitative comparison between \methodname and VD-IT, the strongest baseline on this benchmark, in Figure~\ref{fig:ambiguity}. While many of these examples lack a single ground-truth segmentation, \methodname consistently produces more accurate and coherent masks in the confidently visible regions. For instance, in the first row, our method accurately segments only the clearly visible portions of lava that become apparent after being struck by a wave, whereas VD-IT incorrectly includes the entire wave. In the second row, \methodname reliably segments all regions of glowing water, while VD-IT detects only a few scattered patches. These results demonstrate our model's robustness in ambiguous settings and its capacity to avoid over-segmentation.

\section{Implementation Details}
\label{suppsec:implementation}

\smallsec{Benchmark details and baseline models} We report the details about the benchmarks we used in Table~\ref{supptab:dataset}. 
We quote the results of all the baseline models on Ref-YTB, Ref-DAVIS, and MeViS from their original papers. For the zero-shot evaluation of BURST, VSPW, and Ref-VPS, we report the numbers by running the official checkpoints of MUTR\footnote{\url{https://github.com/OpenGVLab/MUTR}}, UNINEXT\footnote{\url{https://github.com/MasterBin-IIAU/UNINEXT}}, VD-IT\footnote{\url{https://github.com/buxiangzhiren/VD-IT}}, and GLUS\footnote{\url{https://github.com/GLUS-video/GLUS}}.

\begin{table}[t]
    \centering
    \resizebox{\linewidth}{!}{
    \begin{tabular}{l|ccc}
        Benchmark & Type & Training Samples & Testing Samples \\ \hline 
        Ref-COCO~\citep{yu2016modelingcontextreferringexpressions} & Image & 320K & - \\ \hline
        Ref-YTB~\citep{seo2020urvos} & Video & 12,913 & 2,096\\
        MeViS~\citep{ding2023mevis} & Video & 1,712 & 140 \\
        Ref-DAVIS~\citep{khoreva2019video} & Video & - & 90 \\
        BURST~\citep{athar2023burst} & Video & - & 2,049 \\
        VSPW~\citep{miao2021vspw} & Video & - & 343 \\
    \end{tabular}
    }
    \caption{Details about the benchmarks we used for training and evaluation.}
    \label{supptab:dataset}
\end{table}

\smallsec{Training details.} Our approach builds upon two state-of-the-art text-to-video diffusion architectures: ModelScope~\citep{wang2023modelscope} and Wan~\citep{wan2025wan}. Additional video diffusion backbones are evaluated in Section~\ref{suppsec:quantitative}. ModelScope comprises 1.4 billion parameters and extends Stable Diffusion~\citep{blattmann2023stable} with temporal modules. We adopt a two-stage training protocol following~\citet{zhu2024exploring}: in Stage I, we fine-tune only the spatial weights on Ref‐COCO image-text pairs\citep{yu2016modelingcontextreferringexpressions} for one epoch; in Stage II, we fine-tune all network weights for 40 epochs using Ref‐YTB video–text examples~\citep{seo2020urvos} supplemented with 12K Ref‐COCO images converted into pseudo‐videos following \citet{wu2022language}. By contrast, Wan employs a unified diffusion transformer that jointly models spatial and temporal information, without dedicated temporal modules~\citep{peebles2023scalable}. Accordingly, we train this variant in a single stage on the combined Ref‐COCO and Ref‐YTB datasets for 80k steps, with half of the steps trained with images and half trained with videos. Throughout training, the text encoder and VAE remain frozen. 

Unless otherwise stated, all models are trained and evaluated at a resolution of $512\times512$. We use AdamW~\citep{loshchilov2017fixing} for optimization with a constant learning rate of 1e-6. The training batch size is 4 for ModelScope and 8 for Wan, and for each sample, we randomly load an 8-frame video clip for ModelScope and a 17-frame video clip for Wan. We train our model using eight NVIDIA 80GB A100 GPUs, and it takes about 1 week to finish the whole training process. 

For MeViS, we train our MS-1.4B variant by finetuning our Stage I checkpoint jointly on MeViS and Ref-YTB for 37 epochs. We achieve our best results on MeViS by finetuning the Wan-14B checkpoint trained on Ref-COCO and Ref-YTB, for an additional 8 epochs on MeViS.

\smallsec{Evaluation details.} We follow the standard evaluation protocol for Ref-YTB, Ref-DAVIS, and MeViS. For BURST~\citep{athar2023burst} and VSPW~\citep{miao2021vspw}, neither of them contains referring text for the segmented entities. We automatically generate referring expressions using only the category information of the mask entity as ``the $<$class$>$'' (\eg, \emph{the hat}). For VSPW, we conduct our evaluation on the validation set, which has 66 different stuff categories. In the case of BURST, we evaluate the combined validation and test set, which contains 454 classes and a total of 2,049 sequences. 
For inference and evaluation, we follow the standard VSPW protocol for our RVS evaluation. For BURST, we predict masks for all the original frames, and compute the metrics for the annotated ones provided by the dataset. 
For Ref-VPS, to ensure high-quality performances, we perform inference at the original 24 FPS and compute evaluation metrics on the annotated frames at 6 FPS.

\end{document}